\documentclass[10pt]{article} % For LaTeX2e
\usepackage[preprint]{tmlr}

\usepackage{amsmath,amsfonts,bm}

\def\eqref#1{equation~\ref{#1}}
\def\1{\bm{1}}

\DeclareMathAlphabet{\mathsfit}{\encodingdefault}{\sfdefault}{m}{sl}
\SetMathAlphabet{\mathsfit}{bold}{\encodingdefault}{\sfdefault}{bx}{n}

\usepackage{amsmath}
\usepackage{amssymb}
\usepackage{mathtools}
\usepackage{amsthm}

\usepackage{microtype}
\usepackage{hyperref}
\usepackage{url}
\usepackage[capitalize,noabbrev]{cleveref}

\usepackage{graphicx}
\usepackage{subfig}        % provides \subfloat (used in appendix figures)
\usepackage{wrapfig}
\usepackage{float}

\usepackage{booktabs}      % \cmidrule
\usepackage{array}
\usepackage{multirow}      % \multirow
\usepackage{makecell}
\usepackage{colortbl}      % \rowcolor
\def\thickhline{\noalign{\hrule height1pt}}

\usepackage{xcolor}
\definecolor{linkblue}{HTML}{0000EE}
\definecolor{customblue}{HTML}{1E40AF}
\definecolor{boxblue}{RGB}{219,234,254}

\usepackage{tcolorbox}
\tcbuselibrary{skins,breakable}

\usepackage{enumitem}
\setlist[itemize]{leftmargin=5.5mm}

\usepackage{pifont}
\newcommand{\cmark}{\text{\ding{51}}}
\newcommand{\xmark}{\text{\ding{55}}}

\usepackage{algorithm}
\usepackage{algcompatible}

\newcommand{\revision}[1]{\textcolor{black}{#1}}
\colorlet{revisioncolor}{black}

\title{JustLogic: A Comprehensive Benchmark for Evaluating Deductive Reasoning in LLMs}

\author{\name Michael K. Chen
    \email mkc013@ucsd.edu \\
    \addr University of California, San Diego \\
    San Diego, USA
    \AND
    \name Xikun Zhang\thanks{Corresponding author.}
    \email xikun.zhang@rmit.edu.au \\
    \addr Royal Melbourne Institute of Technology \\
    Melbourne, Australia    
    \AND
    \name Yao Li
    \email s4259092@student.rmit.edu.au \\
    \addr Royal Melbourne Institute of Technology \\
    Melbourne, Australia
    \AND
    \name Dacheng Tao
    \email dacheng.tao@gmail.com \\
    \addr Nanyang Technological University \\
    Singapore, Singapore
}

\def\month{10}  % Insert correct month for camera-ready version
\def\year{2026} % Insert correct year for camera-ready version
\def\openreview{\url{https://openreview.net/forum?id=wsDUuu6P38}} % Insert correct link to OpenReview for camera-ready version

\begin{document}

\maketitle

\begin{abstract}
Logical reasoning is a critical component of Large Language Models (LLMs), and substantial research efforts in recent years have aimed to enhance their deductive reasoning abilities. However, existing deductive reasoning benchmarks suffer from significant constraints that restrict their utility, i.e., the lack of task complexity, the presence of prior knowledge as a confounder, and superficial error analysis. To address these deficiencies, we introduce JustLogic, a synthetically generated benchmark for evaluating deduction in propositional logic. Within this fragment, JustLogic is (i) complex, capable of generating a diverse range of linguistic patterns, vocabulary, and argument structures, (ii) prior knowledge independent, substantially reducing the advantage of models possessing prior knowledge so that questions are answerable from the given premises alone, and (iii) capable of in-depth error analysis on the heterogeneous effects of proof size and argument form on model accuracy. Our experimental results on JustLogic reveal that (i) several state-of-the-art (SOTA) reasoning LLMs score above the human average on JustLogic, although only DeepSeek R1's margin is significant on the matched items after correction for multiple comparisons, and (ii) the evaluated non-reasoning models score below the human average. Given the small human study (18 participants, seven questions each), these comparisons are indicative. All code and data are available at \href{https://github.com/michaelchen-lab/JustLogic}{\color{linkblue}{https://github.com/michaelchen-lab/JustLogic}}
\end{abstract}

\section{Introduction}
\label{sec:intro}

Deductive reasoning is a crucial capability for large language models (LLMs). It refers to the process of creating logically valid arguments, where conclusions necessarily follow from the premises, i.e., if an argument's premises are true, the conclusion must also be true. Recent state-of-the-art (SOTA) LLMs \citep{achiam2023gpt, dubey2024llama, jiang2023mistral} have exhibited outstanding performance and consistent improvement across various reasoning benchmarks, including HellaSwag \citep{zellers2019hellaswag}, ARC Challenge \citep{clark2018think} and WinoGrande \citep{sakaguchi2021winogrande}. However, we argue that the existing benchmarks are insufficient and often ineffective for evaluating LLMs' deductive reasoning capabilities.

We identify three major problems with the existing benchmarks. \textbf{First}, they lack complexity, specifically (i) natural language complexity, i.e., how arguments are linguistically expressed, and (ii) argument complexity, i.e., the structure of the argument itself. Manually curated datasets, such as FOLIO \citep{han2022folio} and LogiQA 2.0 \citep{liu2020logiqa, liu2023logiqa} exhibit high natural language complexity but low argument complexity, while synthetic datasets like CLUTRR \citep{sinha2019clutrr} and ProofWriter \citep{tafjord2020proofwriter} exhibit the opposite. \revision{We define these precisely. \emph{Natural-language complexity} is the variety of surface realisations --- grammatical patterns, vocabulary, topics --- used to express an instance; \emph{argument complexity} is the structural variety of the derivation from premises to answer, which for JustLogic is the range of proof sizes and the number of distinct argument structures the generator can produce (Section \ref{subsec:complexity}). Argument complexity is variety \emph{within} a fixed logical language, a different axis from that language's \emph{expressiveness}: JustLogic is propositional throughout, so it is not a claim about covering first-order or quantified reasoning.} Simplicity in either dimension makes these benchmarks prone to overfitting and memorization \citep{mccoy2019right, zhang2022paradox, mirzadeh2025gsm}, thus allowing models to perform well despite underlying weaknesses in logical reasoning. A more detailed analysis can be found in Section \ref{subsec:complexity}.
\textbf{Second}, existing benchmarks often fail to test deductive reasoning in isolation, as models can benefit from prior knowledge. \revision{We measure this with a \emph{prior knowledge independence test}: the question is presented without its premises, so that accuracy above the random baseline indicates that prior knowledge alone suffices (formalised in Section \ref{sec:exp_setup}). Our aim is not that reasoning be divorced from knowledge, but that deductive ability be diagnosable in isolation.} Section \ref{subsec:ci-test-results} shows how prior knowledge can substantially increase accuracy, even in logical reasoning datasets, e.g., FOLIO \revision{(40.0\% with no premises given, against 33.3\% chance)} and LogiQA 2.0 \revision{(52.1\% against 25\%)}. Thus, high accuracy may not reflect strong reasoning capabilities.
\textbf{Third}, many existing benchmarks provide superficial error analysis, leaving key questions unanswered: At what proof size does the model start to fail? How does the model compare to humans at different proof sizes? Which argument forms is the model particularly weak at? These insights are essential for understanding how model performance varies with proof size, longest inference-chain length, and argument form. Section \ref{subsec:error_analysis} demonstrates the importance and usefulness of comprehensive error analysis.

\begin{figure}[H]
	\begin{tcolorbox}[bicolor,colback=gray!10,colbacklower=white,colframe=gray!10,left=1pt,right=1pt]
		\textcolor{customblue}{\textbf{Paragraph:}}
		\begin{itemize}
			\setlength\itemsep{0em}
			\item It is a fact that either relics are artifacts or dogs are capable of barking. [$p \lor q$]
			\item If relics are artifacts, then fertilizers contain phosphorus. [$p \rightarrow r$]
			\item Assuming dogs are capable of barking, we know that fertilizers contain phosphorus. [$q \rightarrow r$]
		\end{itemize}
		\tcblower
        \textcolor{customblue}{\textbf{Statement:}} Fertilizers contain phosphorus. [$r$] \\
		\textcolor{customblue}{\textbf{Question:}} Is the statement true, false, or uncertain? \\
		\textcolor{customblue}{\textbf{Answer:}} True
	\end{tcolorbox}
	\caption{Example of a question adapted using JustLogic's dataset construction algorithm. Note that the statements are \textit{intentionally} factually inaccurate, which we explain and justify in Section \ref{subsec:ds_construct_step2}. Formal notations are included for illustrative purposes and are not provided to models.}
	\label{table:example}
	\end{figure}

% \begin{wrapfigure}{r}{0.6\textwidth}
% 	\begin{tcolorbox}[bicolor,colback=gray!10,colbacklower=white,colframe=gray!10,left=4pt]
% 		\textcolor{customblue}{\textbf{Paragraph:}}
% 		\begin{itemize}
% 			\setlength\itemsep{0em}
% 			\item It is a fact that either relics are artifacts or dogs are capable of barking. [$p \lor q$]
% 			\item If relics are artifacts, then fertilizers contain phosphorus. [$p \rightarrow r$]
% 			\item Assuming dogs are capable of barking, we know that fertilizers contain phosphorus. [$q \rightarrow r$]
% 		\end{itemize}
% 		\tcblower
%         \textcolor{customblue}{\textbf{Statement:}} Fertilizers contain phosphorus. [$r$] \\
% 		\textcolor{customblue}{\textbf{Question:}} Is the statement true, false, or uncertain? \\
% 		\textcolor{customblue}{\textbf{Answer:}} True
% 	\end{tcolorbox}
% 	\caption{Example of a question adapted using JustLogic's dataset construction algorithm. Formal notations are included for illustrative purposes and are not provided to models.}
% 	\label{table:example}
% 	\end{wrapfigure}

Due to these issues, LLMs' deductive reasoning abilities remain ambiguous. In response to the critical need for a reliable benchmark to support ongoing research efforts, we present JustLogic, a novel natural language deductive reasoning benchmark. The task is to determine whether a given statement is true, false, or uncertain, using only the given premises, which are assumed to be true. An example is shown in Figure \ref{table:example}.

% Each instance in JustLogic contains a paragraph of premises and a statement. The task is to determine whether the statement is true, false, or uncertain, based solely on the premises, and assuming they are all true. 

% {\setlength{\extrarowheight}{2pt}
% \begin{table}[ht!]
% 	\def\arraystretch{1.1}
% 	\centering
% 	\caption{Example of a question adapted from the JustLogic train dataset}
% 	\label{table:example}
% 	\begin{tabular}{|p{0.12\linewidth}| p{0.81\linewidth}|}
% 		\hline
% 		\textbf{Paragraph} & \begin{tabular}[c]{@{}l@{}} - Whenever it is true that night blooming plants and trees depend on nectar eating \\ bats for pollination, `if many species are critically endangered, then it is not true \\ that doors are solids' is true. \\ - Night blooming plants and trees depend on nectar eating bats for pollination. \\ - We can assume that many species are critically endangered. \end{tabular}
% 		\\ \hline
% 		\textbf{Statement} & Doors are solids. \\ \hline
% 		\textbf{Answer} & False \\ \hline
% 	\end{tabular}
% \end{table}}
JustLogic’s construction ensures it is (i) complex, (ii) prior knowledge independent, and (iii) capable of in-depth error analysis. \textbf{First}, to achieve high argument and natural language complexity, JustLogic is code-generated rather than manually curated. This allows the generation of a theoretically infinite number of unique argument structures. Natural language sentences are then drawn from GenericsKB-Best \citep{bhakthavatsalam2020genericskb}, a database of 1M+ unique real-world sentences, and inserted into the argument structures, introducing high natural language complexity. \textbf{Second}, since sentences are randomly sampled from the entire GenericsKB-Best dataset, the generated arguments generally do not align with real-world knowledge, thereby substantially reducing prior knowledge as a confounder. \textbf{Finally}, in-depth error analysis is enabled by the programmatic generation process, which enables the inspection of each question's properties, such as proof size and argument form, to investigate their impact on model performance. A comparison between JustLogic and six similar logical reasoning benchmarks is presented in Table \ref{table:comparison}, with further details on dataset construction provided in Section \ref{sec:dataset_construct}.

Using JustLogic, we conducted comprehensive experiments to evaluate the deductive reasoning capabilities of current LLMs. First, our prior knowledge independence test shows that on existing datasets LLMs can answer a substantial proportion of questions without the premises, indicating that accuracy on those benchmarks partly reflects prior knowledge rather than deduction. This is not observed in JustLogic. Second, we benchmarked the performance of SOTA LLMs and human participants using JustLogic. Non-reasoning models and the smaller reasoning models performed below the average human accuracy (73.0\%). GPT-5.5 achieved the highest accuracy (88.8\%), followed by Gemini 2.5 Flash (82.0\%) and DeepSeek R1 (80.9\%). \revision{All three score above the human average, though after correction for multiple comparisons only DeepSeek R1's margin is statistically significant on the 42 items the participants answered, and all three fall short of the highest observed participant accuracy (100.0\%).} Finally, enabled by JustLogic’s code-generated nature, our thorough error analysis examined the heterogeneous impact of argument structure and proof size on model performance. These experiments indicate that JustLogic provides a controlled test of propositional deduction, and that current LLMs still have room for improvement on it.

In summary, our contributions are threefold. First, we evaluate the limitations of existing benchmarks. Second, we introduce the JustLogic benchmark, a synthetic dataset to evaluate deductive reasoning, that addresses the aforementioned limitations. Third, our experiments using JustLogic show that most evaluated models fall below the human average, and that even the strongest model does not reach the highest observed participant accuracy. We posit that LLMs still have room for improvement on propositional deduction as measured by JustLogic, and hope that the JustLogic benchmark will assist researchers in designing and evaluating LLMs.

\begin{table}[H]
	\centering
	\caption{Comparison of JustLogic with other deductive reasoning datasets. The symbol $\sim$ suggests the feature is present but to a limited extent.}
	\label{table:comparison}
		\resizebox{\textwidth}{!}{%
	\begin{tabular}{lcccccc}
		\thickhline & \begin{tabular}[c]{@{}c@{}} High NL\\ Complexity \end{tabular} & \begin{tabular}[c]{@{}c@{}} High Arg.\\Complexity \end{tabular} & \begin{tabular}[c]{@{}c@{}} Prior Knowledge\\ Independence\end{tabular} & \begin{tabular}[c]{@{}c@{}} In-Depth\\ Error Analysis \end{tabular} & \revision{\begin{tabular}[c]{@{}c@{}} Logical\\ Language \end{tabular}} & \revision{\begin{tabular}[c]{@{}c@{}} Task\\ Type \end{tabular}} \\ \thickhline
		CLUTRR & \xmark & \cmark & \cmark & \cmark & \revision{\begin{tabular}[c]{@{}c@{}}Relational\\(family-relation graph)\end{tabular}} & \revision{\begin{tabular}[c]{@{}c@{}}Relation\\prediction\end{tabular}} \\
		ProofWriter & \xmark & \cmark &  \cmark & $\sim$ & \revision{Propositional rules} & \revision{\begin{tabular}[c]{@{}c@{}}Entailment,\\closed-world\end{tabular}} \\
		\revision{ProntoQA-OOD} & \revision{\xmark} & \revision{\cmark} & \revision{\cmark} & \revision{$\sim$} & \revision{\begin{tabular}[c]{@{}c@{}}Propositional /\\simple quantification\end{tabular}} & \revision{Proof generation} \\
		LogiQA 2.0 & \cmark & \xmark & \xmark & $\sim$ & \revision{\begin{tabular}[c]{@{}c@{}}Informal,\\no formal system\end{tabular}} & \revision{Multiple choice} \\
		FOLIO & \cmark & \xmark & \xmark &  $\sim$ & \revision{First-order} & \revision{Entailment} \\
		\revision{BeliefBank} & \revision{\xmark} & \revision{\xmark} & \revision{\xmark} & \revision{$\sim$} & \revision{\begin{tabular}[c]{@{}c@{}}Propositional constraints\\(ConceptNet)\end{tabular}} & \revision{\begin{tabular}[c]{@{}c@{}}Entailment,\\True/False\end{tabular}} \\ \hline
		\textbf{JustLogic} & \cmark & \cmark & \cmark & \cmark & \revision{\textbf{Propositional}} & \revision{\textbf{\begin{tabular}[c]{@{}c@{}}Entailment, open-world\\(True/False/Uncertain)\end{tabular}}} \\ \thickhline
	\end{tabular}}
\end{table}

\section{Related Work}

\subsection{Existing reasoning datasets for Large Language Models}

Reasoning benchmarks are a vital part of LLM evaluation. Some benchmarks measure deductive reasoning in conjunction with natural language inference (NLI), inductive reasoning, and commonsense knowledge: HellaSwag \citep{zellers2019hellaswag} tasks machines to select the most likely follow-up of an event description, WinoGrande \citep{sakaguchi2021winogrande} is a pronoun resolution task, and MuSR \citep{sprague2023musr} tasks machines to solve fictional problems, such as murder mysteries. Other benchmarks measure reasoning on domain knowledge: AI2 Reasoning Challenge (ARC) \citep{yadav2019quick} contains grade-school science questions, while Massive Multitask Language Understanding (MMLU) \citep{hendrycks2020measuring} contains questions across 57 subjects in STEM, humanities, and more. Finally, math-specific benchmarks include GSM-8K \citep{cobbe2021training} and DROP \citep{dua2019drop}.

The aforementioned benchmarks explicitly evaluate skills beyond reasoning and do not specifically define the type of reasoning involved, e.g. inductive, deductive, and analogical. As such, benchmarks that solely test for deductive reasoning have seen a considerable increase in interest. They can be classified into two broad categories: synthetic and manually curated. Synthetically-generated datasets include (i) CLUTRR \citep{sinha2019clutrr}, where a machine must infer the relationship of two family members based on stories, (ii) ProofWriter \citep{tafjord2020proofwriter}, where a machine must deduce a statement's truth value based on a set of facts and rules, and (iii) ProntoQA-OOD \citep{saparov2024testing}, where a machine must prove a statement based on a set of facts. Manually curated datasets include (i) LogiQA 2.0 \citep{liu2023logiqa}, containing manually-translated logical reasoning questions from the Chinese Civil Service Exam, (ii) FOLIO \citep{han2022folio}, containing questions with manually-annotated content using Wikipedia pages, and (iii) ReClor \citep{yu2020reclor}, containing reading comprehension questions from GMAT and LSAT.

\revision{A related line of work constrains model beliefs rather than generating arguments. BeliefBank \citep{kassner2021beliefbank} checks whether a model's True/False answers about an entity are consistent with a bank of constraints drawn from ConceptNet. It differs from JustLogic in three respects: its labels propagate from real-world facts, so its truth values are grounded in world knowledge, whereas JustLogic decouples them by construction; its constraints are single implications and mutual exclusivities, giving single-step instances, whereas JustLogic composes seven argument forms to a controllable proof size; and it is a True/False task, whereas JustLogic includes Uncertain under the open-world assumption. The two are complementary: BeliefBank measures the consistency of knowledge a model already holds, JustLogic measures deduction in isolation from it.}

As discussed earlier, synthetic datasets are prior knowledge independent and exhibit high argument and low natural language complexity; manually curated datasets are the opposite. JustLogic, being synthetic, retains the argument complexity and prior knowledge independence of synthetic datasets while offering natural language complexity comparable to manually curated ones, which we further explain in Sections \ref{subsec:complexity} and \ref{subsec:ci-test-results}.

\subsection{Reasoning in Large Language Models}

As LLMs continue to increase in size, their performance on various reasoning-related benchmarks has improved dramatically. For example, in 2024, Gemini Ultra \citep{team2023gemini} achieved 90.0\% on MMLU when the SOTA model in 2020, UnifiedQA 11B \citep{khashabi2020unifiedqa}, achieved a mere 48.9\%. In 2023, GPT-4 achieved 96.4\% on ARC when the SOTA model in 2020, GPT-3 \citep{brown2020language}, achieved 53.2\%.

The advent of prompting techniques played an important role in developing LLMs' reasoning abilities. In-context learning \citep{dong2022survey} provides LLMs with instructions and examples in the input prompt to guide its response. Chain-of-thought prompting \citep{wei2022chain} prompts LLMs to generate a series of intermediate reasoning steps before arriving at the final answer. Self-consistency decoding \citep{wang2022self} chooses the most consistent answer after sampling multiple chain-of-thought outputs. Least-to-most prompting \citep{zhou2022least} decomposes a complex problem into simpler subproblems, which are then solved sequentially.

As mentioned above, LLMs are conventionally tested on datasets that combine reasoning with other skills. Moreover, existing logical reasoning-specific datasets possess major limitations that call into question the reliability of their evaluations. In response, JustLogic aims to robustly and accurately evaluate the deductive reasoning abilities of LLMs.

\section{Dataset Construction}
\label{sec:dataset_construct}

JustLogic is a programmatically generated dataset designed to evaluate a model's ability of deductive reasoning, specifically its capability to form logically valid arguments. A logically valid argument is one where the conclusion necessarily follows from the premise(s); in other words, given the premises are true, the conclusion must also be true.

We state the scope of the benchmark explicitly. JustLogic evaluates \emph{entailment checking in propositional logic} under the open-world assumption: given a set of premises and a query statement, the task is to decide whether the premises entail the query, entail its negation, or neither. It does not cover first-order logic, quantified reasoning, relational reasoning over structured knowledge, \revision{defeasible inference,} or discourse-level reasoning\revision{, nor reasoning that requires world knowledge}. Claims made in this paper about deductive reasoning ability should be read as applying to this fragment\revision{, and our claims about complexity, prior knowledge independence and error analysis are claims about this fragment as well}.

In order to test this, JustLogic presents a model with a paragraph consisting of premises, followed by a query statement. Based solely on the premises and assuming they are all true, the model needs to determine whether the query statement is true, false, or uncertain. In line with the open-world assumption, the ``Uncertain'' answer refers to cases where the premises neither support nor contradict the query statement.

\begin{figure}[H]
	\centering
	\includegraphics[width=1.0\textwidth]{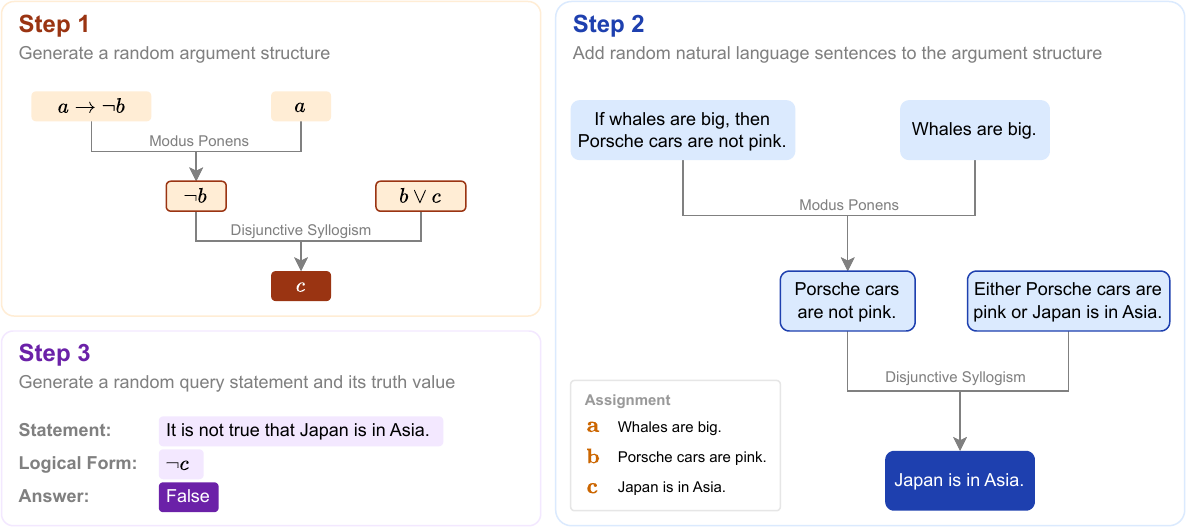}
	\caption{A step-by-step example of how an instance with a proof size of 2 is constructed.}
	\label{fig:dataset_construct}
\end{figure}

The following outlines the process for generating each instance in the dataset, which is exemplified by Figure~\ref{fig:dataset_construct}.

\begin{enumerate}[topsep=0pt]
    \setlength\itemsep{-0.3em}
	\item Step 1: Generate an argument structure.
	\item Step 2: Add natural language statements to the argument structure.
	\item Step 3: Generate a query statement.
\end{enumerate}

\subsection{Step 1: Generate argument structure}

Argument structures are composed of one or more valid argument forms, derived from propositional logic; argument forms are made up of a series of logical forms, defined as symbolic representations of statements. Specifically, the argument forms in our dataset are constructed with the following four logical forms: basic ($x$), negation ($\neg x$), conditional ($x \rightarrow y$), and disjunction ($x \vee y$). \revision{Each of these forms has a clear and non-overlapping surface realisation in English, which is why Table \ref{table:basic_forms} assigns each its own set of expression templates; this, rather than logical minimality, is our criterion for inclusion.} We explicitly define the seven fundamental forms \citep{johnson2006}: modus ponens (MP),  modus tollens (MT), hypothetical syllogism (HS), disjunctive syllogism (DS), reductio ad absurdum (RAA), constructive dilemma (CD), and disjunctive elimination (DE). Table \ref{table:arg_forms} in Appendix \ref{app:arg_forms} provides their formal notations and examples.

The generation process involves several random sampling steps with the following distributions: logical form conclusions are sampled with probabilities 0.8 (Statement), 0.1 (Conditional), and 0.1 (Disjunction); negation is applied to the conclusion with probability 0.3; and argument forms are sampled uniformly at random from the set of applicable forms for each conclusion type. The algorithm to create an argument structure (Appendix \ref{app:alg}) accepts the proof size as input. It first generates a random conclusion and an argument form to support it, which in Figure \ref{fig:dataset_construct} is $c$ and disjunctive syllogism. If the intended proof size has not been reached, one or more premises will become subconclusions, which are supported by new, randomly generated argument forms. In Figure \ref{fig:dataset_construct}, this is exemplified by $\neg b$ becoming a subconclusion, supported by a modus ponens argument. This process continues until the desired proof size is achieved.

Specifically, proof size is defined as the total number of argument forms in the proof structure, rather than the length of the longest inference chain. This definition captures the overall structural complexity of the argument, as each additional argument form introduces new premises and logical dependencies that the model must process. While a proof-size-$d$ instance may contain inference chains of varying lengths, the total number of argument forms provides a comprehensive measure of the reasoning burden imposed on the model. \revision{Proof size and the longest inference chain are not interchangeable: 58.8\% of instances have a chain strictly shorter than their proof size, and at proof size 7 only 21 instances (2.1\%) are genuine seven-step chains. Their joint distribution, and accuracy against both axes, are given in Appendix \ref{app:size_vs_chain}.}

\subsection{Step 2: Adding natural language}
\label{subsec:ds_construct_step2}

Next, the symbolic statements are converted into natural language. Each statement consists of one or more logical forms, \textit{i.e.} variable, negation, conditional, and disjunction. In natural language, these forms can be expressed in a variety of ways. For example, a conditional can be expressed as both ``If $x$, then $y$.'' and ``Given that $x$, $y$ is true.'', where variables $x$ and $y$ are simple propositions. To emulate the diversity of natural language, a list of expressions for each logical form is created using human feedback, and potentially aided by LLMs. Table \ref{table:basic_forms} shows the formal notation of each form, alongside a sample expression and the total number of unique expressions. The variable(s) within each expression are ultimately replaced by randomly selected generic, real-world sentences from GenericsKB-Best \citep{bhakthavatsalam2020genericskb}. The open-source GenericsKB-Best dataset is chosen for its vast collection of simple propositions (1,020,868 sentences) without conditionals, disjunctions, etc. A complete example can be found in Step 2 of Figure \ref{fig:dataset_construct}.

\begin{table}[H]
	\def\arraystretch{1.1}
	\centering
	\caption{Expressions of logical forms.}
	\label{table:basic_forms}
	\begin{tabular}{lllc}
		\thickhline
		& Formal Notation & Sample Expression                        & No. of Expr. \\ \thickhline
		Basic       & $x$            & The claim that $x$ holds true.              & 16                        \\
		Negation    & $\neg x$            & The claim that $x$ does not reflect reality. & 15                        \\
		Conditional & $x \rightarrow y$         & Once we know that $x$, we also know that $y$. & 11                        \\
		Disjunction & $x \vee y$            & It is a fact that either $x$ or $y$.          & 8                         \\ \thickhline
	\end{tabular}
\end{table}

Notably, as shown in Figure 2, the statements are generally factually inaccurate despite being drawn from real-world data. This is intentional. Real-world propositions allow us to generate sentences with diverse grammatical structures that closely emulate human-written arguments. However, factually accurate arguments enable models to bypass deductive reasoning with their prior real-world knowledge, which is experimentally demonstrated in Section \ref{subsec:ci-test-results}. By using real-world yet factually inaccurate statements, we combine realism and prior knowledge independence. 

There are potential concerns that factually inaccurate statements and ``unnatural'' synthetic language may confuse models and lead to artificially low performance. Appendix \ref{app:factual-accuracy} and \ref{app:language_unnaturalness} empirically refute these concerns.

\subsection{Step 3: Generate query statement}

The LLM's task is to determine whether the given query statement is true, false, or uncertain based on the premises provided. Using Figure \ref{fig:dataset_construct} as an example, if we assign the query statement to be the negation of the conclusion, i.e. ``It is not true that Japan is in Asia'', then the answer is false. If the query statement is the same as the conclusion, then the answer is true. If the query statement is unrelated to the premises, then the answer is uncertain. To generate Uncertain instances, a sentence sharing the same topic term as the conclusion is sampled from GenericsKB-Best. The sampler excludes the conclusion and every other sentence already used in the argument, comparing normalised strings so that records differing only in whitespace or terminal punctuation are treated as identical. Appendix~\ref{app:verification} reports an audit of the released data.

% Each paragraph is randomly assigned an answer: True, False, or Uncertain. If the assigned answer is True, the query statement is simply the conclusion of the argument. If the answer is False, the query statement is the negation of the conclusion. If the answer is Uncertain, the query statement is a random sentence from GenericsKB-Best that matches the topic of the conclusion but is unrelated to the argument. Figure \ref{fig:dataset_construct} shows an example of a query statement where the answer is False.

\subsection{Dataset Complexity}
\label{subsec:complexity}

In the context of deductive reasoning datasets, complexity is defined as the variety and comprehensiveness of instances. It can be further divided into two dimensions: natural language complexity and argument complexity, defined formally in Section \ref{sec:intro}. In this section, we highlight the significance of both aspects and how JustLogic compares against other logical reasoning datasets.

\begin{table}[H]
	\centering
	\caption{Statistics of dataset complexity. The structural-scale measure is benchmark-specific; JustLogic uses proof size, which counts all argument forms in the proof structure and is distinct from longest inference-chain length.}
	\label{table:complexity}
	\begin{tabular}{lcccc}
		\thickhline
		& \multicolumn{2}{c|}{Natural Language}                              & \multicolumn{2}{c}{Argument}                                 \\
		& Reading Difficulty $\uparrow$ & \multicolumn{1}{c|}{Vocab. (Domains)} & Structural Scale & Arg. Structures \\ \thickhline
		CLUTRR      & 6.67                & \multicolumn{1}{c|}{1396 (1)}            & 1 - $\infty$                  & $\infty$                     \\
		ProofWriter & 0.96               &  \multicolumn{1}{c|}{101 (×)} & 1 - $\infty$            & $\infty$                     \\
		LogiQA 2.0      & 17.10 & \multicolumn{1}{c|}{10004 (\textgreater{}10)} & ×               & ×                      \\
		FOLIO       & 18.75 & \multicolumn{1}{c|}{4351 (\textgreater{}10)} & 1 - 7             & 76                   \\ \hline
		\textbf{JustLogic}   & \textbf{20.55} & \multicolumn{1}{c|}{\textbf{10557 (\textgreater{}10)}} & \textbf{1 - $\infty$}              & \textbf{$\infty$}                       \\ \thickhline
	\end{tabular}
\end{table}

\textbf{Natural language complexity.} Human language is complex. Statements and arguments of similar meanings can be presented in a variety of ways. Therefore, it is insufficient for models to reason solely with symbols, \textit{e.g.} $x$ and $y$, and basic natural language sentences, \textit{e.g.} “Some birds are yellow.”; they must be capable of reasoning with real-world vocabulary and diverse sentence structures to be useful in practical contexts. 

We measure natural language complexity with (i) reading difficulty, as measured by the Flesch-Kincaid Grade Level test \citep{kincaid1975derivation}, and (ii) lexical diversity, as measured by vocabulary \& domain size. \revision{For (i), the score is presented as a U.S. grade level; the higher the score, the harder the text is to read. Scores greater than 12 should be used to compare the relative difficulty between benchmarks, with higher scores indicating relatively greater textual complexity.} A domain is defined as any topic of interest, such as golf, computers, or traveling; the number of domains is quantified by counting the number of distinct topic categories present in the dataset. For JustLogic, domains are derived from the term 'categories' in GenericsKB-Best, which covers more than 10 broad topic areas. Vocabulary size refers to the number of unique words in the dataset. Given the difficulty of quantitatively capturing linguistic complexity, Appendix \ref{app:nl-complexity-sample} also shows text samples of each benchmark, representative of their complexity.

As shown in Table \ref{table:complexity}, existing synthetic datasets have low natural language complexity, while human-written datasets, such as FOLIO and LogiQA 2.0, exhibit significantly higher complexity. This is expected since synthetic datasets translate symbols in formal logic into natural language using limited templates of sentence structures and vocabulary lists. For example, in ProofWriter, a typical sentence follows the format ``All dogs are (not) red.''. The linguistic patterns of human-written datasets, in contrast, are bound only by human creativity. Despite being synthetic, JustLogic, achieves natural language complexity on par with manually curated datasets, due to its comprehensive selection of expressions and the use of GenericsKB-Best as the source of sentences.

\textbf{Argument complexity.} Argument complexity refers to the diversity of argument structures used in the dataset. A sufficiently high argument complexity is important because humans use a range of argument forms to reason, beyond just conditionals and modus ponens. Moreover, a real-world argument is typically composed of multiple argument forms, due to their inherent complexity. A dataset’s argument complexity is evaluated based on (i) the range of its benchmark-specific structural-scale measure and (ii) the number of unique argument structures. The upper limit of both metrics is calculated based on the theoretical maximum without any additional human input, rather than the highest value used in existing experiments. For example, CLUTRR can generate arbitrary relation lengths, although its experiments used lengths of up to 10; its theoretical upper limit is unbounded. 

Table \ref{table:complexity} shows that synthetic datasets, such as CLUTRR, ProofWriter, and JustLogic, excel in this area, as their generators support arbitrarily large structural scales and numbers of argument structures. Manually curated datasets, in contrast, either lack an explicit structural-scale measure and inventory of argument structures (e.g. LogiQA 2.0), or have finite observed values for both (e.g. FOLIO). While manual datasets require significant human efforts and investment to expand their complexity, synthetic ones can scale trivially. In summary, within propositional logic, JustLogic combines the argument complexity of synthetic datasets with the natural language complexity of manually curated ones.

\subsection{Future-proofing JustLogic}
\label{subsec:futureproof}

As the reasoning abilities of LLMs continue to improve, we expect LLMs to solve the existing JustLogic dataset eventually. To maintain JustLogic's relevance, we leverage its synthetic nature to increase complexity with minimal human input. Argument complexity can be adjusted by increasing the (i) range of proof size \revision{(empirically validated in Appendix \ref{app:futureproof})} and (ii) number of distinct argument forms to \textgreater 7 \revision{(demonstrated for conjunction in Appendix \ref{app:conjunction})}. Natural language complexity can be adjusted by (i) increasing the number of expressions for each logical form and (ii) integrating more complex knowledge bases than GenericsKB. Importantly, these changes are programmatically achievable with minimal man-hours.

Importantly, JustLogic can also effectively tackle benchmark leakage \citep{xu2024benchmarking}, whereby test sets are unintentionally included in LLMs' pretraining data, thus artificially inflating their performance through memorization. Should JustLogic's test set be leaked, a new test set of similar difficulty can be trivially generated, thereby mitigating this problem.

\section{Experimental Setup}
\label{sec:exp_setup}

We first investigate the influence of prior knowledge on evaluating deductive reasoning with JustLogic and other existing benchmarks using our prior knowledge independence test. Next, several SOTA LLMs of various sizes are evaluated using JustLogic. Finally, an in-depth error analysis of the LLMs' results is conducted.

JustLogic contains 7000 instances, equally split amongst proof sizes ranging from 1 to 7. The test set (15\% of all instances) is used for evaluation. Note that the number of instances and range of proof sizes can be easily adjusted using JustLogic's open-source dataset generation program.

\subsection{Prior Knowledge Independence Test}

The task for deductive reasoning benchmarks is typically framed as $CQO \rightarrow A$: Given a context $C$, consisting of $n$ premises ($P = \{p_1,p_2,...,p_n\}$), a question $Q$, and $m$ answer options ($O = \{o_1,o_2,...,o_m\}$), determine the correct answer $A$. To assess the influence of prior knowledge on determining answer $A$, the prior knowledge independence test is framed as $QO \rightarrow A$. No context $C$ is provided, and the prompt instructs the LLM to answer the question based on prior knowledge alone. A few-shot prompt with examples is provided to guide the model's response format, but no contextual premises are included. The full prompt is shown in Appendix~\ref{app:ci-test}.

If prior knowledge is not useful, the LLM should be unable to answer question $Q$ without $C$, and the accuracy for the prior knowledge independence test should approximate random probability $\frac{1}{m}$. Benchmarks exhibiting such accuracies are deemed prior knowledge independent.

While context ablation has been used in NLP to assess model reliance on input context, we systematically apply it across five deductive reasoning benchmarks to quantify the influence of prior knowledge on benchmark accuracy. While any LLM will be suitable, we use GPT-4 for its extensive prior knowledge. Similar results are replicated using Llama3-70B in Appendix \ref{app:ci-test-additional}. The test is conducted on JustLogic and other logical reasoning benchmarks, i.e. CLUTRR, ProofWriter, LogiQA 2.0, and FOLIO.
% \vspace{-2mm}

\subsection{Evaluating LLMs' Deductive Reasoning}

Our task, as illustrated in Figure \ref{table:example}, follows the conventional formulation: $CQO \rightarrow A$. Question $Q$ is ``Is the statement $S$ true, false, or uncertain?''; there are 3 answer options, where $O = \{\text{true},\text{false},\text{uncertain}\}$. Prompts begin with a preamble, providing (i) the task instructions, (ii) a list of argument forms in propositional logic, and (iii) the available answer options.

\revision{We categorize models into reasoning and non-reasoning types based on their training methodology. Reasoning models are those explicitly trained with chain-of-thought reasoning via reinforcement learning, including DeepSeek R1, GPT-5.5, OpenAI o1, and Gemini 2.5 Flash, while non-reasoning models are standard instruction-tuned models without such training, including GPT-4o and Llama3. We evaluate reasoning and non-reasoning models of various sizes: Llama3-8B-Instruct \citep{dubey2024llama}, Llama3-70B-Instruct, GPT-4o-mini-2024-07-18, GPT-4o-2024-05-13, DeepSeek R1 Distill Qwen 14B, DeepSeek R1 \citep{guo2025deepseek}, OpenAI o1-mini-2024-07-18, OpenAI o1-2024-12-17 \citep{jaech2024openai}, Gemini 2.5 Flash \citep{comanici2025gemini25}, and GPT-5.5-2026-04-23, using zero-shot, few-shot, and CoT prompting methods (with more tested in Appendix \ref{app:prompt_performance}). Due to limited model access, 70 samples are used for OpenAI o1 and 350 for DeepSeek R1, OpenAI o1-mini, and Qwen 14B (R1 Distill). GPT-5.5 and Gemini 2.5 Flash are evaluated on the full test set of 1,050 samples. To ensure fairness, the selected subset has the same proportion of each proof size as the entire test set. Further implementation details are provided in Appendix \ref{app:exp-details}. Human performance is also measured; the experiment settings are in Appendix \ref{app:human}.}

We perform an error analysis to examine the heterogeneous effects of argument form and proof size on model accuracy. Accuracy for each argument form is measured using questions with a proof size of 1, because those with a proof size of \textgreater 1 typically have \textgreater 1 argument forms. \revision{Since the generator retains the full proof structure of every instance, we additionally analyse composed proofs, holding proof size fixed and testing the effect of each form's presence with a Cochran--Mantel--Haenszel test; results are reported in Section \ref{subsec:error_analysis}. Accuracy within each gold class, and the associated prediction bias, are reported in Appendix \ref{app:per_class}.}
\section{Results}

\subsection{Prior Knowledge Independence Test}
\label{subsec:ci-test-results}

The results of JustLogic and four other benchmarks are shown in Table \ref{table:ci_test_results}; note that a benchmark is considered more prior knowledge independent when the absolute difference $|\Delta|$ between model accuracy and random probability is smaller, regardless of whether accuracy falls above or below random chance. A larger $|\Delta|$ in either direction indicates that prior knowledge is influencing model responses. Thus, the smaller the $|\Delta|$, the better. The $|\Delta|$s of CLUTRR and ProofWriter are relatively low, while those of LogiQA 2.0 and FOLIO are nontrivially higher. This is because the former are synthetic datasets, while the latter are manually curated. When a question is code-generated, it generally bears no correlation with reality, e.g. ``Is it true, false, or uncertain that Gary is not red.'' from ProofWriter. Such questions are only answerable by reasoning over the context $C$. LogiQA 2.0 and FOLIO, on the other hand, often contain questions that are answerable purely using background knowledge, such as ``Did the United States won the most medals in the last summer Olympic games?'' from FOLIO. We posit that this is an unintentional consequence of the human bias to align the question's truth value with reality. While human curation enhances the question's realism, it compromises the test for deductive reasoning.

\revision{The JustLogic benchmark's $|\Delta|$ is 0.4\%, the lowest of the five benchmarks, and 2.7\% when the equal-proportion instruction is removed from the prompt (Appendix \ref{app:ci-test-additional}) --- still far below FOLIO (9.6\%) and LogiQA 2.0 (25.4\%), though CLUTRR (0.7\%) is then lower.} The reason for this is twofold: first, JustLogic is a synthetic dataset, thus eliminating the human bias. Second, while JustLogic uses real-world statements, their truth value is nonetheless randomly determined. For example, the statement ``doors are solids'' is factually true. However, by deducing from the paragraph, the correct answer is ``False''. Thus, using prior knowledge for many questions is not only unhelpful but also meaningfully decreases accuracy. Appendix \ref{app:ci-test-additional} repeats this test with Llama3-70B-Instruct, where JustLogic's $|\Delta|$ is 5.7\% and remains substantially lower than that of the manually curated benchmarks\revision{, and also reports the test with the equal-proportion instruction removed from the prompt: $|\Delta|$ moves by at most 3.5 points and in both directions, with one of the ten model--benchmark comparisons significant after Holm correction}.

\addtolength{\tabcolsep}{-3pt}
\begin{table}[H]
	\centering
	\caption{Results of Prior Knowledge Independence Test. \textbf{The lower the $|\Delta|$, the better.}}
	\label{table:ci_test_results}
	\begin{tabular}{lccc}
		\thickhline
		& $|\Delta| \downarrow$ & Accuracy (\%) & Random (\%) \\ \thickhline
		CLUTRR & 2.0 & 8.3 & 6.3 \\
		ProofWriter & 3.7 & 37.0 & 33.3 \\
		LogiQA 2.0 & 27.1 & 52.1 & 25.0 \\
		FOLIO & 6.7 & 40.0 & 33.3 \\ \hline
		JustLogic & \textbf{0.4} & 33.7 & 33.3 \\ \thickhline
	\end{tabular}
\end{table}
\addtolength{\tabcolsep}{3pt}

\subsection{Evaluating LLMs' Deductive Reasoning}
\label{subsec:llm-eval}

As shown in Table~\ref{table:results_all}, the best-performing model is GPT-5.5 (88.8\%), followed by Gemini 2.5 Flash (82.0\%) and DeepSeek R1 (80.9\%). \revision{On each pair's largest shared subset, GPT-5.5's margins over Gemini 2.5 Flash and DeepSeek R1 remain significant after Holm correction ($p=2.6\times10^{-9}$ and $p=9.0\times10^{-4}$, respectively), whereas the difference between Gemini 2.5 Flash and DeepSeek R1 does not.} OpenAI o1 scores 72.9\% on its 70-item subset; Appendix \ref{app:o1} examines its results on the common subset. Several general observations can be made from these results: first, the difference between Llama3-70B and GPT-4o under CoT prompting is 1.0 percentage point and is not statistically significant (exact paired McNemar test with Holm correction). Second, Llama3-8B using CoT achieved 57.8\% accuracy, compared with 53.1\% for zero-shot Llama3-70B.

% OpenAI o1 is trained to reason with chain-of-thought prompts using a `reinforcement learning algorithm' \citep{openai-o1-2024}. We hypothesize that the use of reinforcement learning on CoT prompting further enhances the deductive reasoning capabilities offered by CoT prompting alone.

\addtolength{\tabcolsep}{-3pt}
\begin{table}[H]
	\centering
	\caption{\revision{Model and Human Evaluation Results, with Wilson 95\% intervals on the CoT column. Sample sizes differ by model and are given in Section \ref{sec:exp_setup}.}}
	\label{table:results_all}
    \color{revisioncolor}
	\begin{tabular}{lcccc}
		\thickhline
		& \textbf{0-shot} & \textbf{Few-shot} & \textbf{CoT} & \revision{\textbf{CoT 95\% CI}} \\ \thickhline
		Random Probability & 33.3 & 33.3 & 33.3 & \revision{-} \\ \hline
		
		Llama3-8B-Instruct & 49.8 & 41.8 & 57.8 & \revision{[54.8, 60.8]} \\
		Llama3-70B-Instruct & 53.1 & 57.8 & 64.6 & \revision{[61.6, 67.4]} \\
		GPT-4o-mini & 53.0 & 54.7 & 51.8 & \revision{[48.8, 54.8]} \\
		GPT-4o & 53.8 & 58.3 & \textbf{65.6} & \revision{[62.7, 68.4]} \\ \hline
        
        OpenAI o1-mini & - & - & 62.0 & \revision{[56.8, 66.9]} \\
		OpenAI o1 & - & - & 72.9 & \revision{[61.5, 81.9]} \\
        Qwen 14B (R1 Distill) & - & - & 61.7 & \revision{[56.5, 66.7]} \\
        DeepSeek R1 & - & - & 80.9 & \revision{[76.4, 84.6]} \\
        Gemini 2.5 Flash & - & - & 82.0 & \revision{[79.6, 84.2]} \\
        GPT-5.5 & - & - & \textbf{88.8} & \revision{[86.7, 90.5]} \\\hline
		
		Human Average & 73.0 & 73.0 & 73.0 & \revision{[64.7, 80.0]} \\
		Highest Observed Participant Accuracy & 100.0 & 100.0 & 100.0 & \revision{-} \\ \thickhline
	\end{tabular}
\end{table}
\addtolength{\tabcolsep}{-3pt}

\revision{Because the evaluation subsets are nested, every model can also be re-scored on the 70 items OpenAI o1 was run on; Appendix \ref{app:common_subset} reports that view. The ordering is not stable across the two: on the common subset DeepSeek R1 moves above Gemini 2.5 Flash.}
% The specific versions are GPT-4-0613, GPT-4o-2024-05-13, OpenAI o1-mini-2024-09-12, OpenAI o1-2024-12-17

Human performance (73.0\%) is higher than the observed accuracies of all non-reasoning models and most reasoning models. Four of the eighteen participants answered all seven assigned questions correctly; a 7/7 score has a Wilson 95\% interval of [64.6, 100.0]. On the 42 items answered by participants, paired bootstrap comparisons with Holm correction found a statistically significant margin over the human average only for DeepSeek R1 (adjusted $p=0.014$; GPT-5.5: $p=0.070$; Gemini 2.5 Flash: $p=0.343$). Notably, the rapid improvement of LLMs on JustLogic, from 65.6\% (GPT-4o) to 88.8\% (GPT-5.5), raises the question of whether the benchmark is approaching saturation. However, since JustLogic is grounded in symbolic propositional logic with deterministic correct answers, we expect LLMs to eventually achieve 100\% accuracy, much like how calculators perform arithmetic accurately every time. Rather than indicating saturation, this trajectory reflects genuine progress in deductive reasoning, and JustLogic's difficulty can be readily scaled as models improve (see Appendix~\ref{app:futureproof}). The human evaluation results may not be fully reflective of the range of human abilities across different ages, educational levels, and areas of expertise, especially given the small sample size, and should be interpreted as indicative guidelines rather than conclusive measurements.

\subsection{Error Analysis}
\label{subsec:error_analysis}

\begin{figure}[H]
	\centering
    \subfloat{{\includegraphics[width=7.9cm, keepaspectratio]{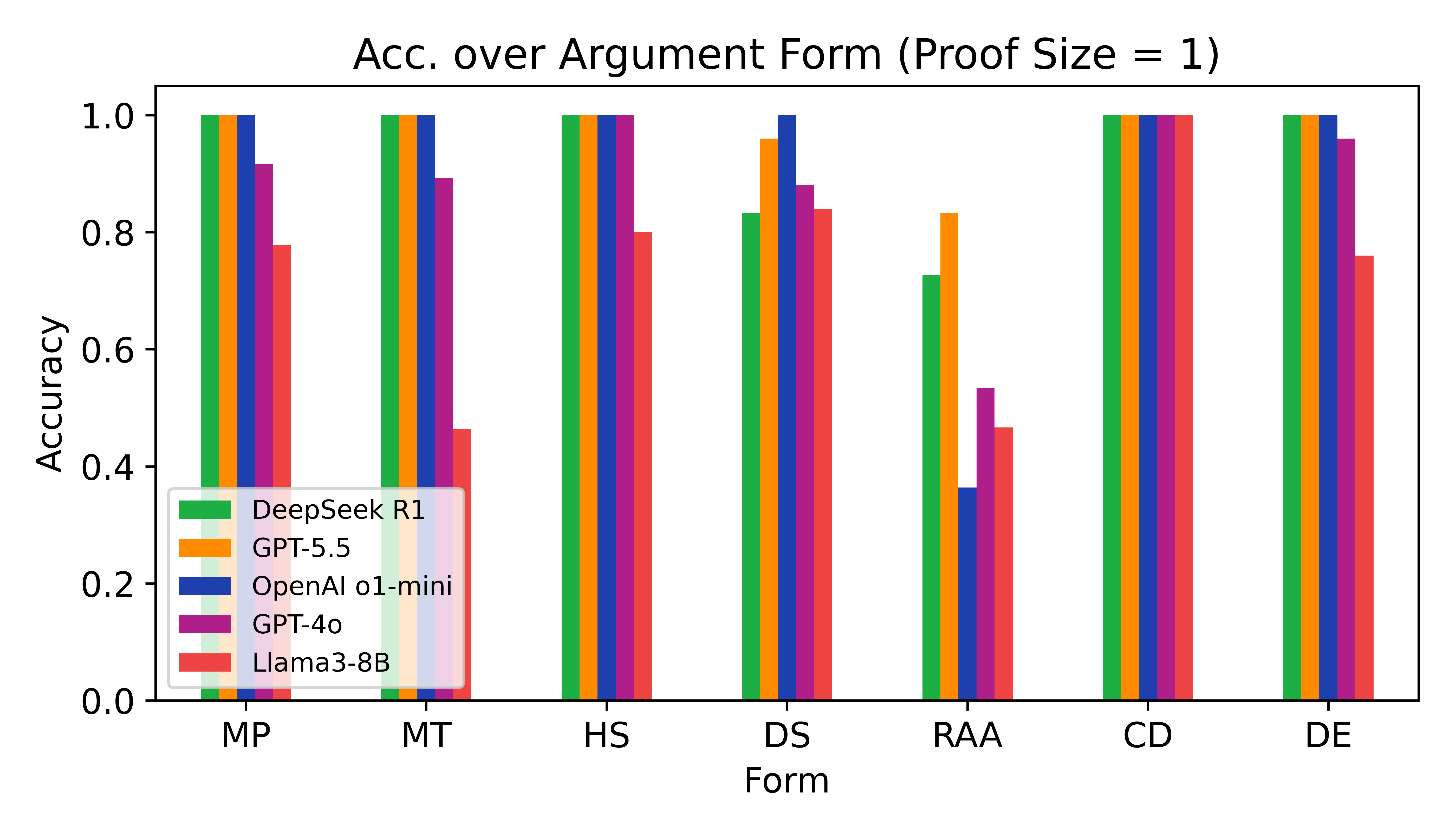} }}
	\quad
	\subfloat{{\includegraphics[width=7.9cm, keepaspectratio]{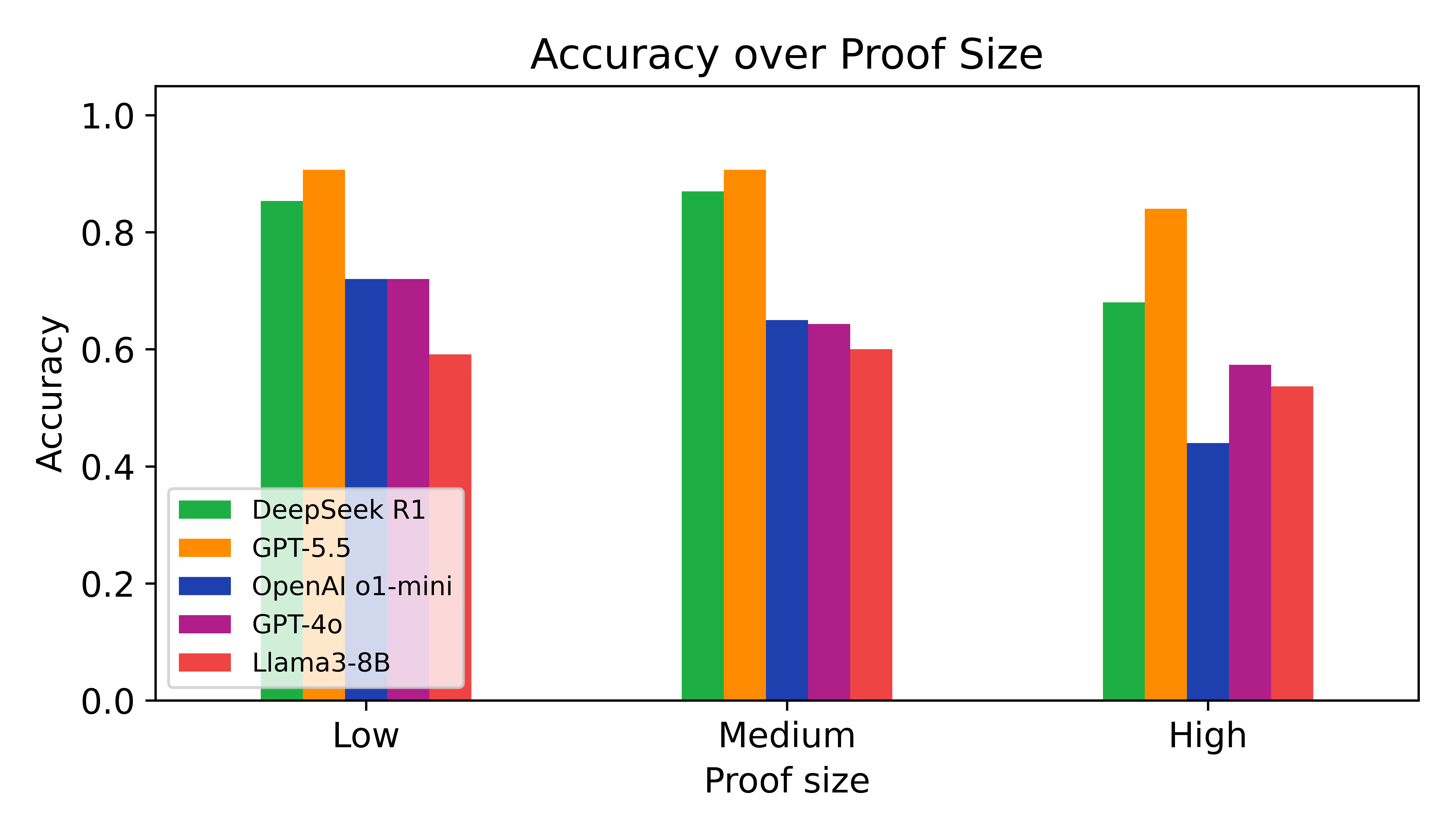} }}
	\caption{\revision{How argument form and proof size affect accuracy for various models. At proof size 1 the number of instances per form is uneven --- constructive dilemma has one instance per model and hypothetical syllogism two to five --- so the left panel should be read qualitatively.}}
	\label{fig:acc_depth_and_arg_form}%
\end{figure}

\revision{Figure \ref{fig:acc_depth_and_arg_form} illustrates the model accuracy by argument form (left) and proof size (right). The best models of their respective categories are presented: Llama3-8B (small, non-reasoning model), GPT-4o (large, non-reasoning model), OpenAI o1-mini (small, reasoning model), and DeepSeek R1 and GPT-5.5 (large, reasoning models). To mitigate noise, proof sizes are grouped into low (1–3), medium (4–5), and high (6–7) categories. The qualitative analysis of failure modes can be found in Appendix \ref{app:failure-cases}.}

% \caption{\revision{How argument form\protect\footnotemark \:  and reasoning depth affects accuracy for various models.}}

While the relative accuracies of argument forms are heterogeneous across models, some forms perform distinctly better than others. For example, HS and CD achieve considerably higher performance than MT, DS, and RAA. Interestingly, the former argument forms are more commonly used by humans than the latter ones, which potentially hints at the cause of this observation.

\revision{This ordering is not confined to proof size 1. Holding proof size fixed and testing each form's presence with a Cochran--Mantel--Haenszel test over all 7,000 instances, reductio ad absurdum ($-10.4$ points, macro-averaged over models) and modus tollens ($-8.0$) carry consistent negative effects, while modus ponens ($+5.7$) and constructive dilemma ($+6.1$) carry positive ones. The ordering observed at proof size 1 survives composition: the Spearman correlation between a form's proof-size-1 accuracy and its presence effect in composed proofs is positive for \revision{all nine models on which the test is possible (OpenAI o1's 70-item subset is too small), with a median $\rho$ of 0.51}.}

\revision{As for proof size, model accuracies generally decrease as proof size increases, consistent with our intuition. Notably, GPT-5.5 achieves the highest accuracy across all proof sizes, maintaining strong performance even at large proof sizes (0.84), while OpenAI o1-mini sees a sharp decline in performance once proof size increases. In fact, at medium and large proof sizes, o1-mini no longer performs better than non-reasoning models, i.e. GPT-4o and Llama3-8B. These observations show that the evaluated models differ in their robustness to larger proof structures, with GPT-5.5 maintaining the highest observed accuracy across the proof-size groups. Since proof size and longest inference-chain length are strongly but imperfectly correlated, Appendix \ref{app:size_vs_chain} reports complementary analyses along both axes.}

% \footnotetext{MP = Modus Ponens, MT = Modus Tollens, HS = Hypothetical Syllogism, DS = Disjunctive Syllogism, RAA = Reductio Ad Absurdum, CD = Constructive Dilemma, DE = Disjunction Elimination}

% On the other hand, recent literature \citep{zhou2024larger} shows that difficult problems for humans are not necessarily difficult for LLMs, and vice versa. The relationship between reasoning depth and LLM accuracy calls for further investigation.

% On the one hand, the fluctuations on all three trendlines are likely a result of small sample sizes: OpenAI o1-preview has 6 samples per depth, while the other 2 have 150. We expect the trend to be more explicit with a larger number of samples. On the other hand, this phenomenon corroborates the recent findings \cite{zhou2024larger} that reveal questions difficult for human may not be difficult for LLMs, and vice versa. \vspace{-2mm}

\section{Conclusion}

Deductive reasoning is an important component of real-world reasoning, alongside background knowledge and instruction following. To advance SOTA LLM capabilities, researchers must diagnose whether their models are limited by their deductive reasoning ability, which must be evaluated in isolation. Against the backdrop of existing reasoning benchmarks that are confounded by the use of prior knowledge and other factors, JustLogic is designed to serve this diagnostic function for propositional deduction by being (i) complex in both argument structure and natural language, (ii) prior knowledge independent, and (iii) capable of in-depth error analysis. Notably, GPT-5.5 achieved the highest overall accuracy (88.8\%), followed by Gemini 2.5 Flash (82.0\%) and DeepSeek R1 (80.9\%). On the items answered by participants, these models scored above the human average, but only DeepSeek R1's margin was statistically significant after correction for multiple comparisons. We recommend that LLM researchers utilize JustLogic as a key pillar in a broader suite of benchmarks that encompasses the range of skills required for real-world reasoning.

\section{Limitations}
\label{future-works}

\revision{The construction of the Uncertain class leaves two surface regularities that a model can exploit without reasoning. True and False queries are the conclusion or its negation, so their proposition necessarily occurs in the paragraph, while Uncertain queries were drawn from outside the argument and did not; Uncertain queries were also less often rendered through an expression template (37.9\%, against 94.2\% and 99.1\%). A rule that exploits them and performs no reasoning reaches 79.6\% on the test set. Appendix \ref{app:artifacts} quantifies both and reports how much of each model's accuracy they account for.}

\revision{Because the generator is public, a model could also be optimised against its template inventory and sampling distribution rather than against propositional deduction. The regeneration mechanism of Section \ref{subsec:futureproof} mitigates leakage of a particular test set but not this form of targeting.}

While JustLogic already achieves higher or similar natural language complexity to existing deductive reasoning benchmarks, as shown in Section \ref{subsec:complexity}, linguistic complexity can be further enhanced to emulate human-written prose, e.g., news articles and fiction stories. Notably, LLMs can be introduced in Step 2 of JustLogic's dataset construction process, whereby instead of randomly selecting sentences from GenericsKB, an LLM can generate fictional statements and scenarios, e.g. ``John's favorite food is hamburgers.''. While LLM generation has been successful in datasets involving inductive reasoning and commonsense knowledge, e.g., MuSR \citep{sprague2023musr}, it is currently too unreliable for deductive reasoning due to several common mistakes, e.g., ignoring instructions, hallucination, and invalid logic. Nonetheless, as LLMs become more reliable, LLM generation is a promising approach worthy of further exploration.

Error analysis using JustLogic can also be further explored. Interesting research questions include: Are models able to use argument forms appropriately? At which step of the argument chain does the model usually fail? What are the most common reasons for failure? These insights may be useful for fine-tuning models for logical reasoning tasks \citep{liu2023logicot} and model guidance \citep{beurer-kellner24a}.

\subsubsection*{Broader Impact Statement}

\revision{JustLogic measures entailment checking in propositional logic; accuracy on it should not be read as general reasoning ability, and we intend it as one diagnostic within a broader evaluation suite. Because the generator is public, performance could be improved by optimising against it rather than against deduction; should a test set be leaked or targeted, an equivalent one can be regenerated (Section \ref{subsec:futureproof}), though this does not address optimisation toward the generator itself. Sentences are drawn from GenericsKB-Best and inherit its topical, stylistic and cultural tendencies; because labels are determined by argument structure rather than by sentence content, these tendencies do not affect the correctness of any label, but they do shape the vocabulary and subject matter a model is exposed to.}

\revision{The human study was conducted on Amazon Mechanical Turk; participation was voluntary and compensated at \$24 per hour, and the survey posed no foreseeable risks to participants. Responses were collected without identifiers and are reported only in aggregate. With 18 participants answering seven questions each, the human baseline should be read as indicative rather than as a population estimate.}

%\subsubsection*{Author Contributions}
%If you'd like to, you may include a section for author contributions as is done
%in many journals. This is optional and at the discretion of the authors. Only add
%this information once your submission is accepted and deanonymized. 

%\subsubsection*{Acknowledgments}
%Use unnumbered third level headings for the acknowledgments. All
%acknowledgments, including those to funding agencies, go at the end of the paper.
%Only add this information once your submission is accepted and deanonymized. 

\bibliography{tmlr}
\bibliographystyle{tmlr}

\newpage
\appendix

\section{Argument Forms}
\label{app:arg_forms}

\addtolength{\tabcolsep}{6pt}
\begin{table}[H]
	\def\arraystretch{0.9}
	\centering
	\caption{An overview of the argument forms in the JustLogic dataset.}
	\label{table:arg_forms}
	\begin{tabular}{lll}
		\thickhline
		& Formal Notation                                                           & Example \\ \thickhline
		Modus Ponens            & \begin{tabular}[c]{@{}l@{}}$p \rightarrow q$\\ $p$\\ $\vdash q$\end{tabular} & \begin{tabular}[c]{@{}l@{}} If the sky is blue, then the dog is happy.\\ The sky is blue. \\ Therefore, the dog is happy. \end{tabular} \\ \hline
		Modus Tollens           & \begin{tabular}[c]{@{}l@{}}$p \rightarrow q$\\ $\neg q$\\ $\vdash \neg p$\end{tabular}                                                                          & \begin{tabular}[c]{@{}l@{}} If the sky is blue, then the dog is happy.\\ The dog is not happy. \\ Therefore, the sky is not blue. \end{tabular}        \\ \hline
		Hypothetical Syllogism  & \begin{tabular}[c]{@{}l@{}}$p \rightarrow q$\\ $q \rightarrow r$\\ $\vdash p \rightarrow r$\end{tabular} & \begin{tabular}[c]{@{}l@{}} If the sky is blue, then the dog is happy.\\ If the dog is happy, the owner is happy. \\ Therefore, if the sky is blue, the owner is happy. \end{tabular}        \\ \hline
		Disjunctive Syllogism   & \begin{tabular}[c]{@{}l@{}}$p \vee q$\\ $\neg p$\\ $\vdash q$\end{tabular} & \begin{tabular}[c]{@{}l@{}} Either the dog is barking or the dog is asleep. \\ The dog is not barking. \\ Therefore, the dog is asleep. \end{tabular}    \\ \hline
		Reductio ad absurdum    & \begin{tabular}[c]{@{}l@{}}$p \rightarrow q$\\ $p \rightarrow \neg q$\\ $\vdash \neg p$\end{tabular} & \begin{tabular}[c]{@{}l@{}} If the dog is calm, the owner is around. \\ If the dog is calm, the owner is not around. \\ Therefore, the dog is not calm. \end{tabular}        \\ \hline
		Constructive Dilemma    & \begin{tabular}[c]{@{}l@{}}$p \vee q$\\ $p \rightarrow r$\\ $q \rightarrow s$\\ $\vdash r \vee s$\end{tabular} & \begin{tabular}[c]{@{}l@{}} Either the sky is blue or it is raining. \\ If the sky is blue, the race can start. \\ If it is raining, the race is delayed.\\ Therefore, either the race can start or it is delayed. \end{tabular}        \\ \hline
		Disjunction Elimination & \begin{tabular}[c]{@{}l@{}}$p \vee q$\\ $p \rightarrow r$\\ $q \rightarrow r$\\ $\vdash r$\end{tabular} & \begin{tabular}[c]{@{}l@{}} Either the sky is blue or it is raining. \\ If the sky is blue, the dog is cheerful. \\ If it is raining, the dog is cheerful.\\ Therefore, the dog is cheerful. \end{tabular} \\ \thickhline
	\end{tabular}
\end{table}

\section{Algorithm for Step 1 of JustLogic's Dataset Construction}
\label{app:alg}

\begin{algorithm}[H] % ICML usually prefers [tb] (top/bottom) over [H]
\caption{Pseudocode for Step 1 (Generate argument structure) of JustLogic's Dataset Construction Process. It generates the argument structure using level-order construction until the desired number of argument forms $D$ is reached. Note that proof size is defined as the total number of argument forms $D$ (the size of the proof structure), not the length of the longest inference chain.}
\label{alg:step1}
\begin{algorithmic}[1]
    \REQUIRE $D$ \COMMENT{Target number of argument forms}
    \STATE $\phi_0 \gets \textsc{SampleLogicalForm}()$ \COMMENT{Sample a random final conclusion}
    \STATE $a_0 \gets \textsc{SampleArgumentForm}(\phi_0)$ \COMMENT{Samples a random argument form for the final conclusion}
    \STATE $\mathcal{L} \gets [a_0]$
    \STATE $d \gets 1$ \COMMENT{Tracks the no. of argument forms}
    \WHILE{$d < D$}
        \STATE $P \gets \textsc{GetAllPremises}(\mathcal{L})$ \COMMENT{Extract all premises of all $a$ in $\mathcal{L}$}
        \STATE $k \gets \textsc{SampleUniform}(1, \min(D - d, |P|))$ \COMMENT{Sample an integer from 1 to $\min(D-d, |P|)$}
        \STATE $P_{\text{sub}} \gets \textsc{SampleRandomSubset}(P, k)$ \COMMENT{Select $k$ premises to become subconclusions}
        \STATE $\mathcal{L}' \gets [\,]$
        \FORALL{$\phi \in P_{\text{sub}}$}
            \STATE $a \gets \textsc{SampleArgumentForm}(\phi)$
            \STATE Append $a$ to $\mathcal{L}'$
        \ENDFOR
        \STATE $\mathcal{L} \gets \mathcal{L}'$
        \STATE $d \gets d + |P_{\text{sub}}|$
    \ENDWHILE
    \STATE \textbf{return} Argument structure rooted at $\phi_0$
\end{algorithmic}
\end{algorithm}

\section{Proof Size and Longest Inference Chain}
\label{app:size_vs_chain}

\revision{Proof size counts every argument form in the proof structure, while the longest inference chain counts only the longest root-to-leaf path through it. Table \ref{table:size_vs_chain} summarises how the two diverge, over all 7,000 instances and recovered from the released proof structures. They coincide up to proof size 2 and separate steadily thereafter: at proof size 7 the chain ranges from 3 to 7 with a mean of 4.67, and only 21 instances (2.1\%) are genuine seven-step chains. Across the whole dataset 58.8\% of instances have a chain strictly shorter than their proof size, and the two correlate strongly but not perfectly (Pearson 0.888, Spearman 0.901).}

\begin{table}[H]
	\centering
	\caption{\revision{How the longest inference chain is distributed within each proof size. There are 1,000 instances at each proof size.}}
	\label{table:size_vs_chain}
	\revision{
	\begin{tabular}{cccc}
		\thickhline
		\textbf{Proof size} & \begin{tabular}[c]{@{}c@{}}\textbf{Longest chain}\\ \textbf{range}\end{tabular} & \begin{tabular}[c]{@{}c@{}}\textbf{Longest chain}\\ \textbf{mean}\end{tabular} & \begin{tabular}[c]{@{}c@{}}\textbf{Chain $<$ proof size}\\ \textbf{(\%)}\end{tabular} \\ \thickhline
		1 & 1 & 1.00 & 0.0 \\
		2 & 2 & 2.00 & 0.0 \\
		3 & 2--3 & 2.51 & 48.7 \\
		4 & 2--4 & 3.14 & 78.3 \\
		5 & 3--5 & 3.74 & 91.2 \\
		6 & 3--6 & 4.23 & 95.2 \\
		7 & 3--7 & 4.67 & \textbf{97.9} \\ \thickhline
	\end{tabular}}
\end{table}

\begin{figure}[H]
	\centering
	\includegraphics[width=\textwidth, keepaspectratio]{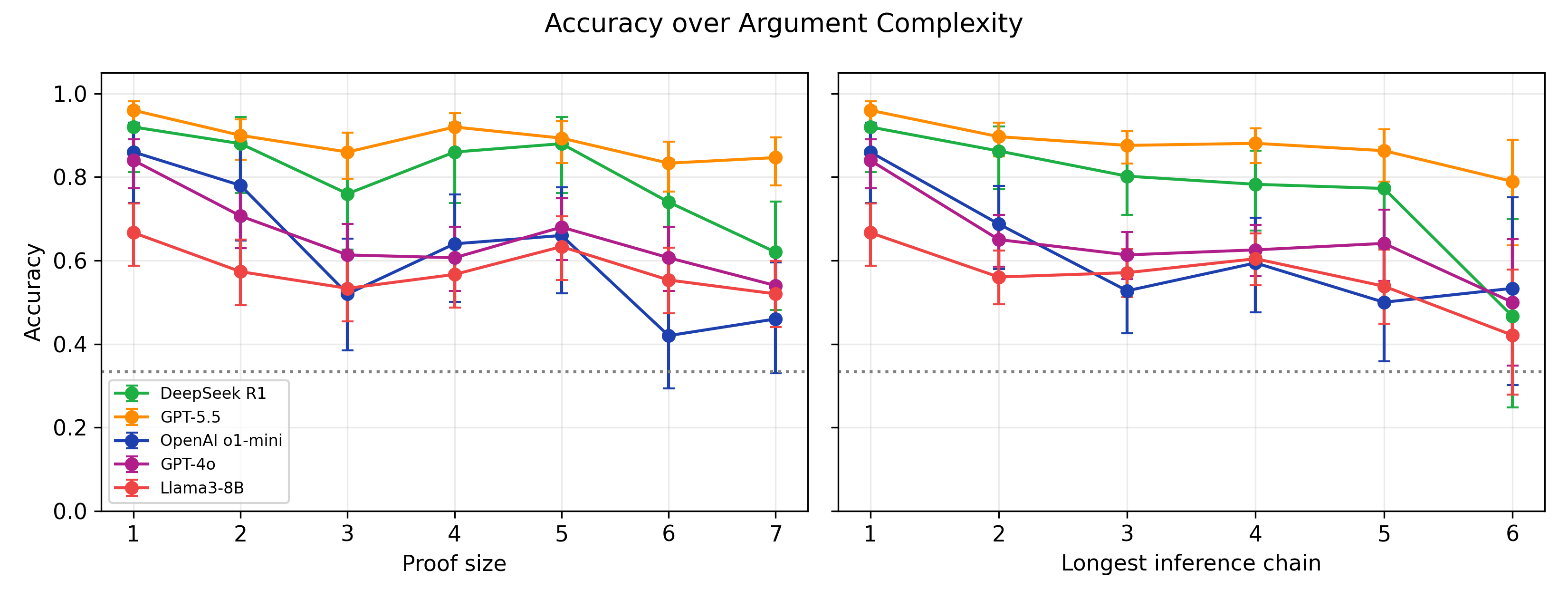}
	\caption{\revision{Accuracy against proof size (left) and the longest inference chain (right). Points with fewer than ten instances are omitted.}}%
	\label{fig:both_metrics}%
\end{figure}

\revision{Figure \ref{fig:both_metrics} reports accuracy against both axes for the models of Figure \ref{fig:acc_depth_and_arg_form}. The qualitative trend is the same under either metric: accuracy falls as the argument grows, and GPT-5.5 stays close to flat while OpenAI o1-mini declines steeply. Holding one metric fixed and varying the other does not separate them at this sample size: of 42 comparisons in which proof size is held fixed and the chain varies, none is significant after Holm correction, and of 36 comparisons in the other direction, one is.}

\section{Sample texts from deductive reasoning benchmarks}
\label{app:nl-complexity-sample}

Beyond metrics like vocabulary size and number of domains, the degree of natural language complexity can be straightforwardly determined by manually inspecting the linguistic patterns of a given benchmark. Table \ref{table:sample-texts} shows sample texts from CLUTRR, ProofWriter, ProntoQA-OOD, SimpleLogic, LogiQA 2.0, FOLIO, and JustLogic.

% \makecell[c]{CLUTRR \\ \citep{sinha2019clutrr}}

\begin{table}[H]
\centering
\def\arraystretch{1.1}
\caption{Sample texts from various deductive reasoning benchmarks.}
\label{table:sample-texts}
\begin{tabular}{>
{\centering\arraybackslash}p{0.17\linewidth}p{0.70\linewidth}}
\thickhline
\multicolumn{1}{c}{\textbf{Benchmark}} & 
  \multicolumn{1}{c}{\textbf{Sample Text}}
\\ \thickhline
CLUTRR \text{\citep{sinha2019clutrr}} & Lorraine and her brother Kevin went to see a movie. Clarence took his granddaughter Lorraine to the movies and they enjoyed themselves. \\ \hline
ProofWriter \text{\citep{tafjord2020proofwriter}} & The bald eagle is not rough. The bear does not need the bald eagle. The dog needs the bear. If someone is rough then they chase the bald eagle. If someone needs the bear then they are not blue... \\ \hline
ProntoQA-OOD \text{\citep{saparov2024testing}} & Lempuses are bitter. Every lempus is a lorpus. Brimpuses are vumpuses. Tumpuses are impuses. Each impus is not hot. Every numpus is a sterpus. Each shumpus is brown. Sterpuses are fast. Every vumpus is not small... \\ \hline
SimpleLogic \text{\cite{zhang2022paradox}} & If messy and hypocritical and lonely, then shiny. If tame, then friendly. If plain and shiny and homely, then nervous. If tender, then hypocritical. If dull and impatient and plain, then tame. If spotless, then perfect. If elegant and tender, then homely... \\ \hline
LogiQA 2.0 \text{\citep{liu2023logiqa}} & In the past 10 years, the sales of personal notebook computers of a computer company have continued to grow, but the growth rate is lower than the growth rate of the company's total sales of all products. \\ \hline
FOLIO \text{\citep{han2022folio}} & All people who regularly drink coffee are dependent on caffeine. People regularly drink coffee, or they don't want to be addicted to caffeine, or both. No one who doesn't want to be addicted to caffeine is unaware that caffeine is a drug... \\ \hline
\rowcolor{boxblue}
JustLogic & Provided that sound travels through different kinds of matter, we know that every soul is a candidate for immortality. It is a common misconception that every soul is a candidate for immortality. \\ \thickhline
\end{tabular}
\end{table}

Evidently, JustLogic exhibits significantly greater natural language complexity than CLUTRR, ProofWriter, ProntoQA-OOD, and SimpleLogic, because the latter benchmarks programmatically generate every sentence, while JustLogic extracts its sentences from GenericsKB, a natural language text database. Thus, the former benchmarks rely on a limited number of grammar templates, reducing their linguistic complexity. JustLogic exhibits similar levels of complexity to FOLIO. LogiQA 2.0 is more complex because it is human-curated and not backed by a formal logic system (unlike how JustLogic is backed by propositional logic). Without a formal logic system, LogiQA 2.0's argument complexity suffers, as shown in Table \ref{table:complexity}, which compromises its ability to evaluate deductive reasoning in LLMs.

\section{Prior Knowledge Independence Test}
\label{app:ci-test}

A sample prompt for the prior knowledge independence test, based on the example in Figure \ref{table:example}, is shown below in Figure \ref{fig:ci_test_example}. Note that the answer options vary depending on the benchmark. For example, the options for LogiQA are A, B, C, and D, while those of CLUTRR are 16 possible family relations.

\begin{figure}[H]
    \begin{tcolorbox}[bicolor,colback=gray!10,colbacklower=white,colframe=gray!10,left=2pt,right=2pt]
        \textcolor{customblue}{\textbf{Instructions:}}
        \begin{itemize}
            \setlength\itemsep{0em}
            \item Use the knowledge you currently have to answer as accurately as possible.
            \item You have 3 answer options: True, False, and Uncertain.
            %\item There should be roughly an equal proportion of each option.
            \item \textit{Add 5-10 examples here}
        \end{itemize}
        \tcblower
        \textcolor{customblue}{\textbf{Question:}} Is the following statement true, false, or uncertain? \\
        \textcolor{customblue}{\textbf{Statement:}} Doors are solids. \\
        \textcolor{customblue}{\textbf{Answer:}} True.
    \end{tcolorbox}
    \caption{Example of a prior knowledge independence test prompt.}
    \label{fig:ci_test_example}
\end{figure}

\section{Experiment Implementation Details}
\label{app:exp-details}

The hyperparameters for the Llama3 models are decided largely based on the recommendations in the original paper \citet{dubey2024llama}, which are as follows: temperature of 0.6, top p of 0.9, presence penalty of 1.15, length penalty of 1. For DeepSeek R1 and Qwen-14B (R1 Distill), the recommended temperature of 0.6 is used. For OpenAI models (GPT-4o, GPT-4o-mini, OpenAI o1, OpenAI o1-mini, and GPT-5.5) and Gemini 2.5 Flash, the default temperature is used. All evaluations are conducted using the OpenAI, OpenRouter, and Google Generative Language APIs, with model costs ranging from \$0.0003 per question for Llama3-8B-Instruct to approximately \$0.26 for GPT-5.5.

With regards to prompting methods, 3-shot prompting is chosen for few-shot experiments because it produces the highest accuracies compared to 6 and 9-shot. Chain-of-thought prompts also contain three examples. In the interest of fairness, all prompting techniques contain similar general instructions, which are as follows:

\begin{tcolorbox}[breakable, colback=boxblue,colframe=boxblue,left=2pt,right=2pt]
You are given a paragraph of facts/premises, followed by a statement. Perform logical reasoning with propositional logic on the paragraph to determine the truth value of the statement. \\

Here is the list of argument forms: 

\begin{itemize}
    \setlength\itemsep{0em}
    \item Modus Ponens
    \item Modus Tollens
    \item Hypothetical Syllogism
    \item Disjunctive Syllogism
    \item Reductio ad absurdum
    \item Constructive Dilemma
    \item Disjunction Elimination
\end{itemize}

You must answer with either one of the 3 options:
\begin{itemize}
    \setlength\itemsep{0em}
    \item TRUE: When the premises in the paragraph lead to the statement
    \item FALSE: When the premises in the paragraph directly contradict the statement
    \item UNCERTAIN: When the premises in the paragraph neither support nor contradict the statement
\end{itemize}

Do not use your prior knowledge; your answer must be solely determined by the information within the paragraph. Assume that all premises in the paragraph are true. \\

Question: Is the statement true, false, or uncertain?
    \end{tcolorbox}

\revision{As for the additional prompting techniques are explored in Appendix \ref{app:prompt_performance}, the tree-of-thought framework contains two prompts at each step: candidate generation and candidate evaluation. In addition to the general instructions above, the candidates generation prompt is shown below.}

\begin{tcolorbox}[breakable, colback=boxblue,colframe=boxblue,left=2pt,right=2pt]
Let's reason step by step. Generate 3 alternative possible next steps, based on the question and the answer so far. Each step consists of a single argument form, e.g. modus ponens. The question takes 1 or more steps to solve. \\

Note that these 3 steps are NOT sequential. They must be alternatives to the same step.
\end{tcolorbox}

\revision{As for candidate evaluation, the prompt is shown below. Note that the model may terminate the exploration prematurely by indicating a final answer. A practical consideration is that models tend to conclude \textit{too early}; the prompt should be designed to emphasize exploration and instruct not to conclude unless sufficiently certain.}

\begin{tcolorbox}[breakable, colback=boxblue,colframe=boxblue,left=2pt,right=2pt]
Of the possible next steps, choose the one that **most directly advances the reasoning process** toward determining the truth value of the statement.  
Select the best next step to continue reasoning toward the answer. Do not conclude with TRUE, FALSE, or UNCERTAIN yet — unless:

\begin{itemize}
    \setlength\itemsep{0em}
    \item All relevant reasoning paths have been explored, and
    \item No further logical deduction is possible or necessary.
\end{itemize}

Otherwise, output only the next reasoning step, using one valid argument form. Your goal is to build a full reasoning chain, not jump to conclusions. \\

Only if this step logically completes the reasoning chain and no further analysis is needed, then conclude with one of: TRUE, FALSE, or UNCERTAIN.
\end{tcolorbox}

\section{Additional Experimental Validations of the JustLogic Benchmark}

\subsection{Prior Knowledge Independence Test using Other Models}
\label{app:ci-test-additional}

\revision{To ensure that the results of the prior knowledge independence test, conducted with GPT-4 in Section \ref{subsec:ci-test-results}, are replicable, we conduct the same test using Llama3-70B-Instruct. The results are shown in Table \ref{table:ci_test_results_llama70b}. The results confirm that JustLogic maintains a high degree of prior knowledge independence compared to human-curated benchmarks such as FOLIO and LogiQA 2.0. However, we note that JustLogic's $|\Delta|$ differs between GPT-4 (0.4\%) and Llama3-70B (5.7\%). This discrepancy is likely attributable to differences in instruction-following ability between the two models: Llama3-70B may be more susceptible to leveraging prior knowledge despite explicit instructions not to do so, resulting in a higher $|\Delta|$. Nonetheless, JustLogic's $|\Delta|$ remains substantially lower than that of LogiQA 2.0 (23.3\%) and FOLIO (10.0\%) under Llama3-70B, confirming that JustLogic is more prior knowledge independent than manually curated benchmarks regardless of the model used for testing. Interestingly, ProofWriter's accuracy is significantly lower than random, which is potentially problematic since models may be biased \textit{against} statements whose truth-value aligns with reality.}

\addtolength{\tabcolsep}{-3pt}
\begin{table}[H]
	\def\arraystretch{1.2}
	\centering
	\caption{Results of Prior Knowledge Independence Test using Llama3-70B-Instruct. \textbf{The lower the $|\Delta|$, the better.}}
	\label{table:ci_test_results_llama70b}
    \color{revisioncolor}
	\begin{tabular}{lccc}
		\thickhline
		& $|\Delta| \downarrow$ & Accuracy (\%) & Random (\%) \\ \thickhline
		CLUTRR & 5.4 & 11.7 & 6.3 \\
		ProofWriter & 8.6 & 24.7 & 33.3 \\
		LogiQA 2.0 & 23.3 & 48.3 & 25.0 \\
		FOLIO & 10.0 & 43.3 & 33.3 \\ \hline
		JustLogic & \textbf{5.7} & 39.0 & 33.3 \\ \thickhline
	\end{tabular}
\end{table}
\addtolength{\tabcolsep}{3pt}

\revision{Figure \ref{fig:ci_test_example} shows the prompt without the instruction ``There should be roughly an equal proportion of each option.'', which the prompts used for Tables \ref{table:ci_test_results} and \ref{table:ci_test_results_llama70b} did contain. We therefore repeat the test with the instruction removed, on pinned gpt-4-0613 and llama-3.3-70b-instruct snapshots and on the same items, so that the effect of the prompt is not confounded with two years of model drift. Re-running the original prompt on those snapshots reproduces the published values to within 1.4 points, except on LogiQA 2.0, where the 1.8-point difference reflects the stricter answer parser used in the re-run; the two prompts are therefore directly comparable. Table \ref{table:debiased} reports $|\Delta|$ with the instruction removed. Relative to the matched original-prompt run it moves by at most 3.5 points and in both directions, and of the ten model--benchmark comparisons only one is significant after Holm correction (Llama3-70B on LogiQA 2.0, exact McNemar on paired items). The largest single change is JustLogic under GPT-4, whose $|\Delta|$ rises to 2.7.}

\begin{table}[H]
	\centering
	\caption{\revision{Prior Knowledge Independence Test with the equal-proportion instruction removed from the prompt. }}
	\label{table:debiased}
	\revision{
	\begin{tabular}{lcc}
		\thickhline
		& GPT-4 & Llama3-70B-Instruct \\ \thickhline
		CLUTRR & \textbf{0.7} & 9.7 \\
		ProofWriter & 2.7 & 4.6 \\
		LogiQA 2.0 & 25.4 & 25.0 \\
		FOLIO & 9.6 & 8.1 \\ \hline
		JustLogic & 2.7 & \textbf{4.4} \\ \thickhline
	\end{tabular}}
\end{table}

\subsection{Impact of Factual Accuracy on Model Performance}
\label{app:factual-accuracy}

Given that JustLogic randomly chooses sentences from GenericsKB to add to each instance's argument structure, the final conclusion may be factually accurate or inaccurate in the real world. For example, if the conclusion is ``It is not true that Japan is in Asia.'', then the conclusion is factually inaccurate. Thus, there is a concern that models underperform due to confusion arising from factually inaccurate conclusions. Moreover, since some conclusions are factually accurate, such instances may exhibit artificially high performance.

To study these concerns, we conducted the following empirical study. If the above concerns are true, we expect factually inaccurate conclusions to perform worse than factually accurate ones. Because all GenericsKB sentences are factually accurate, we can straightforwardly deduce each conclusion's factual accuracy. For example, $x \vee y$ is factually accurate while $\neg x$ is not.

\begin{figure}[H]
	\centering
	\subfloat{{\includegraphics[width=8cm, keepaspectratio]{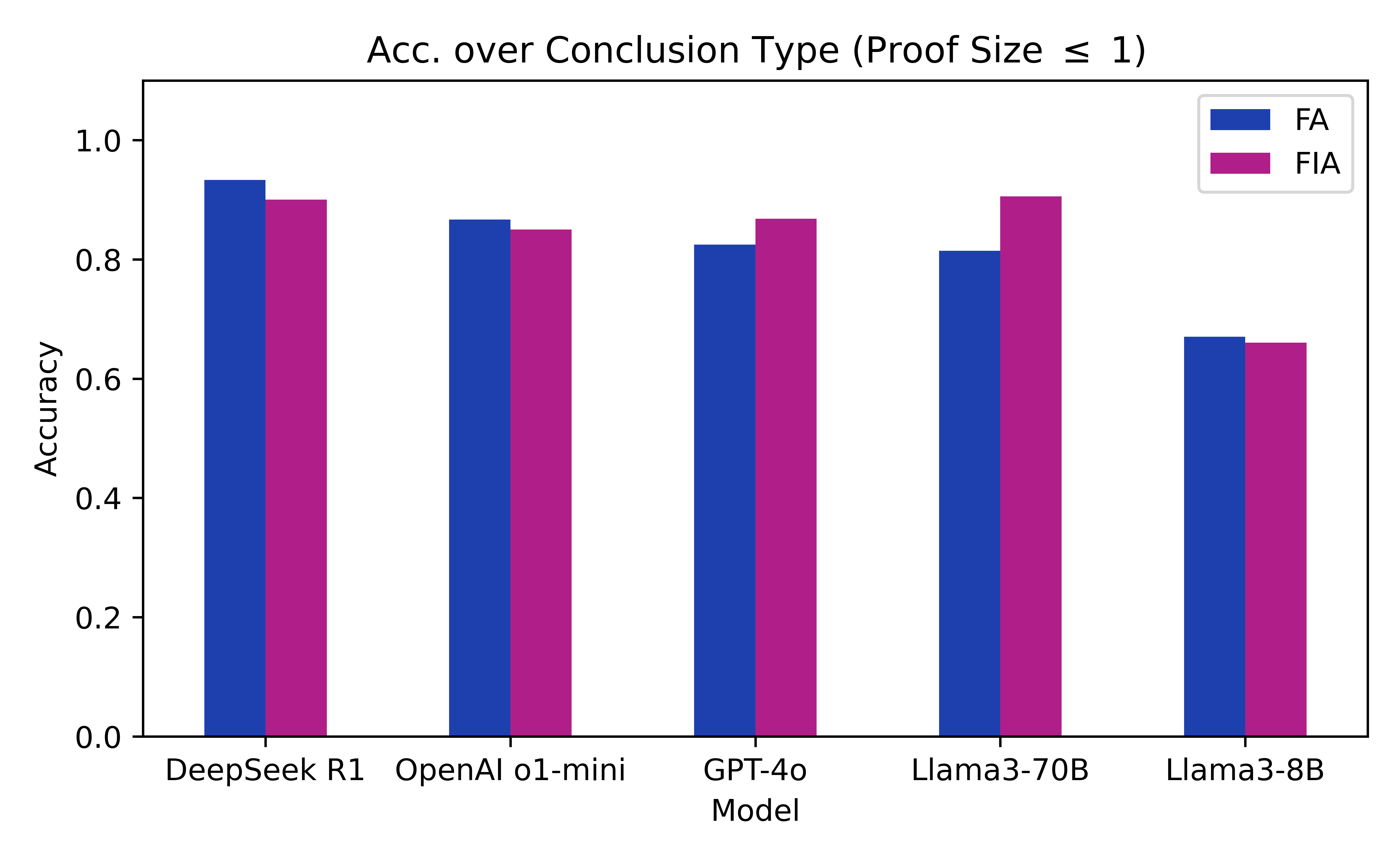} }}%
    \;
	\subfloat{{\includegraphics[width=8cm, keepaspectratio]{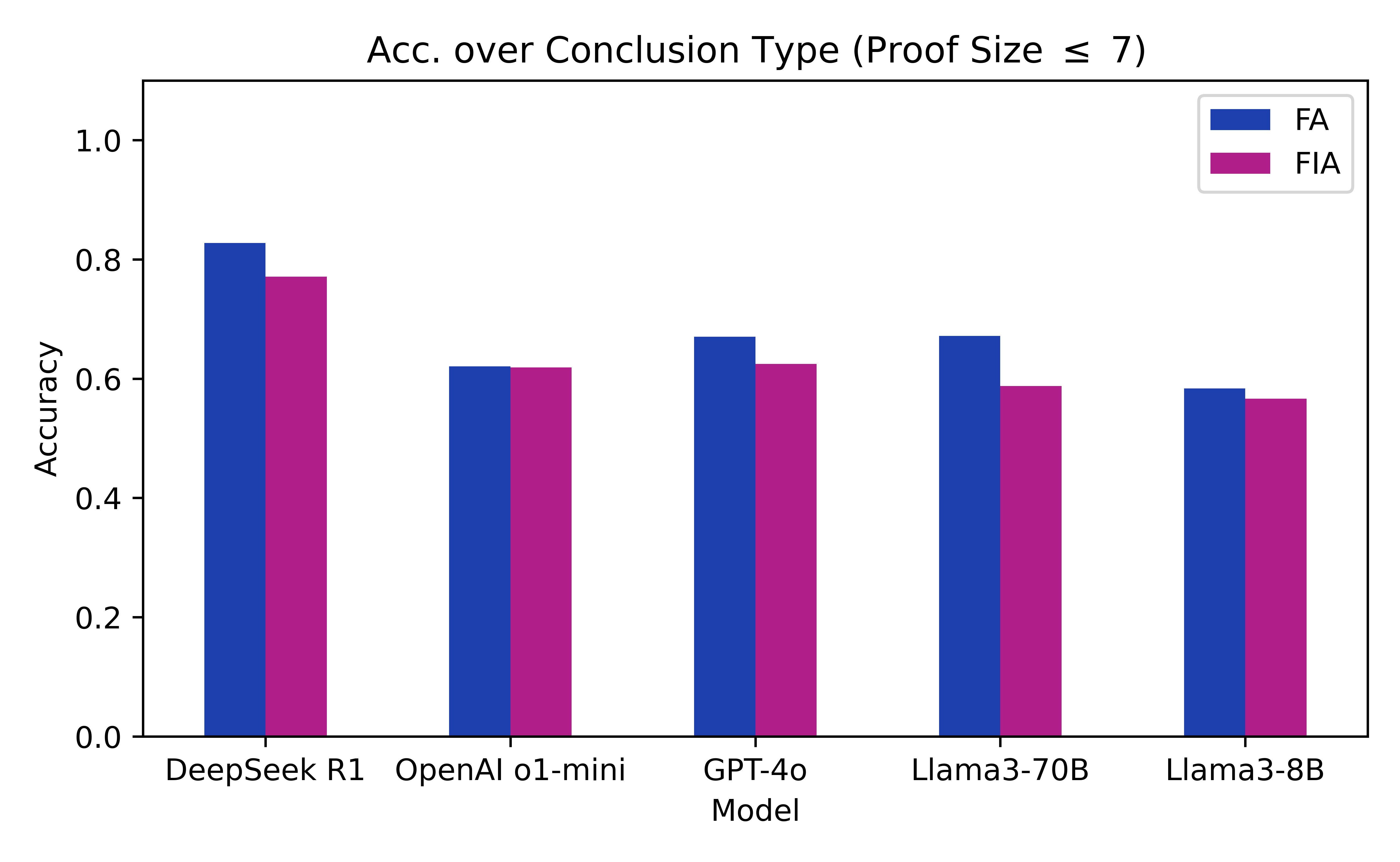} }}
	\caption{How factual accuracy of conclusions affects model accuracy.}%
	\label{fig:acc_conc_type}%
\end{figure}

Figure \ref{fig:acc_conc_type} shows the comparison of accuracies for five models: DeepSeek R1, OpenAI o1-mini, GPT-4o, Llama3-70B, and Llama3-8B; the left represents when proof size is 1, while the right represents when proof size is 7 or less.

These results reject the hypothesis that factually inaccurate conclusions perform worse than factually accurate ones; there is no consistent trend between both conclusion types. In fact, when proof size=1, factually inaccurate conclusions exhibit higher performance for some models! At proof sizes of 7 or less, GPT-4o and Llama3-70B saw a decrease in \textit{relative} accuracy of factually inaccurate statements, DeepSeek R1 and Llama3-8B maintained similar accuracies, while OpenAI o1-mini saw an improvement.

There are two reasons for these results. First, our prompt explicitly instructs models to answer the question only using the paragraph provided and without using prior knowledge. The full prompt is shown in Appendix \ref{app:exp-details}. Moreover, in few-shot prompts, the examples provided include conclusions where their factual accuracy does not match the correct answer. These measures encourage models to ignore prior knowledge and answer questions without considering the factual accuracy of conclusions in the real world.

Second, how LLMs treat factual accuracy when reasoning deductively depends on the LLM's training: specifically, the model's ability to follow prompt instructions to ignore prior knowledge. For example, DeepSeek R1 biases toward factually inaccurate conclusions when deductively reasoning, while OpenAI o1-mini exhibits little difference in performance. Should an LLM exhibit significant differences in performance between factually accurate and inaccurate conclusions, it suggests the LLM has room for improvement in instruction following.

Importantly, the ability to deduce whether premises lead to a conclusion without using prior knowledge is a fundamental human skill: we use it to evaluate whether a debater’s speech or journalist’s article supports their position. The inclusion of both factually accurate and inaccurate instances in JustLogic is a feature, not a bug.

\subsection{Impact of Language ``Unnaturalness'' on Model Performance}
\label{app:language_unnaturalness}

\revision{Given that JustLogic is synthetically generated, there is a concern that its natural language may be highly unnatural to models, potentially hindering their ability to reason deductively. To study this concern, we compare the model perplexity of JustLogic, two other human-curated benchmarks (FOLIO and LogiQA), and two other synthetic benchmarks (CLUTRR and ProofWriter). Llama3-8B-Instruct (a non-reasoning model) and DeepSeek R1 Distill Qwen 14B (a reasoning model) are used. If JustLogic's language is indeed highly unnatural, we expect its model perplexity to be significantly higher than other benchmarks.}

\begin{figure}[H]
	\centering
	\includegraphics[width=8cm, keepaspectratio]{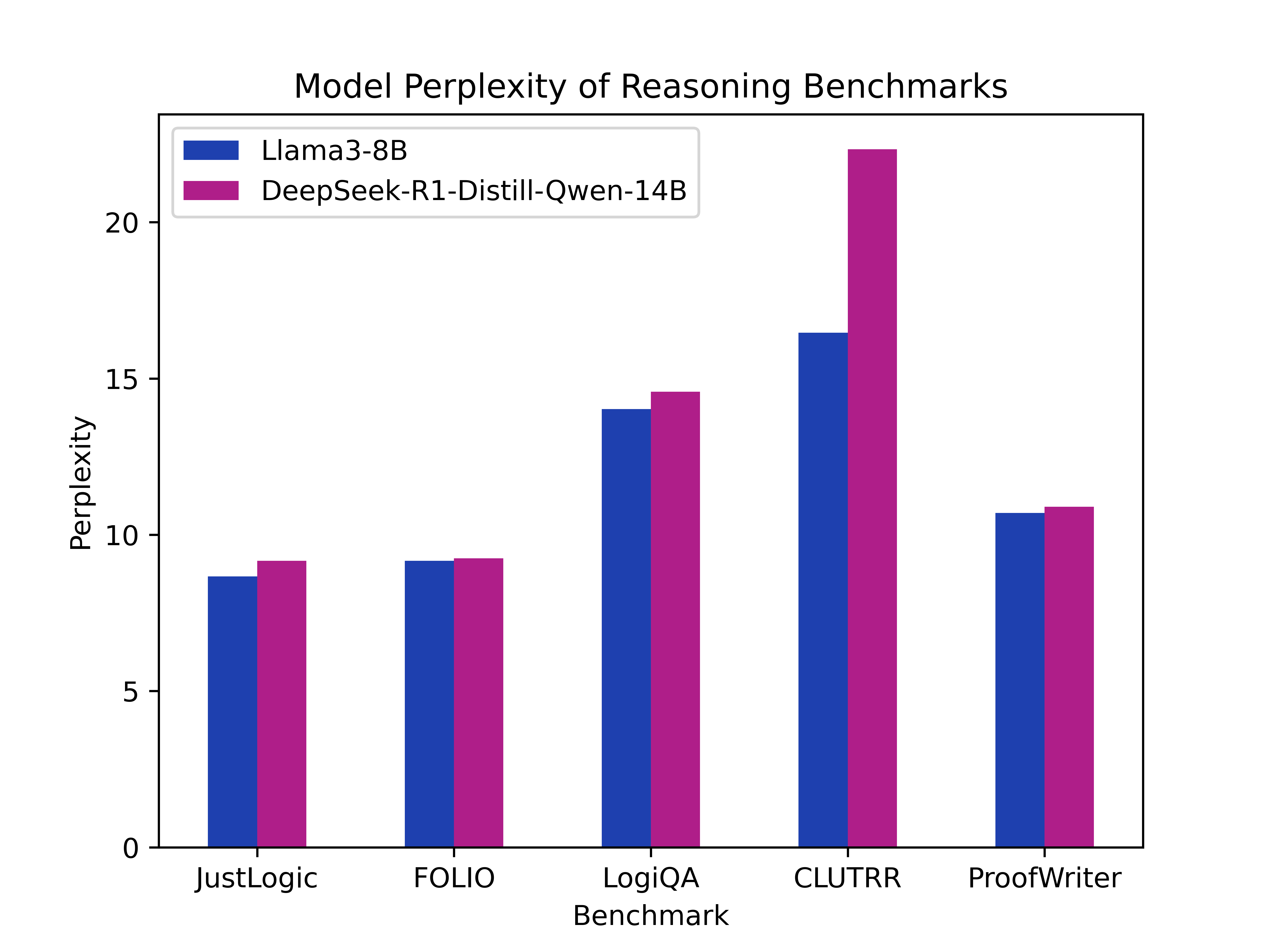}
	\caption{\revision{Model perplexities of various logical reasoning benchmarks.}}%
	\label{fig:benchmark_perplexity}%
\end{figure}

\revision{The results, as shown in Figure \ref{fig:benchmark_perplexity}, reject the aforementioned hypothesis. JustLogic's model perplexities are comparable to FOLIO and lower than the rest. This shows that despite JustLogic's higher linguistic complexity (Table \ref{table:complexity}), its syntactic patterns are well understood by models. CLUTRR's and ProofWriter's higher perplexities are likely due to their unnatural symbolic-like language; LogiQA's higher perplexities are likely because its questions are originally in Chinese and did not shed their foreign syntactic patterns when translated into English. Examples can be found in Appendix \ref{app:nl-complexity-sample}.}

\revision{Therefore, while JustLogic's natural language may seem unnatural to human readers, their syntactic patterns are highly intuitive to LLMs compared to other reasoning benchmarks. JustLogic's language likely does not hinder LLMs' understanding of the questions.}

\subsection{Symbolic Verification of Benchmark Correctness}
\label{app:verification}

\revision{Every released instance includes its complete nested proof tree, so the benchmark can be audited independently of the generator: we re-parse each tree and decide entailment by exhaustive truth-table enumeration, which is exact rather than approximate. Across all 7,000 instances of the train, validation and test splits, every tree parses and its recovered leaf premises match the released premise list; the premise set is satisfiable and entails the stated conclusion in every case, and no True instance is derivable from a proper subset of its premises. All 27,000 argument-form applications are locally valid, with no invalid step in any of the seven forms: modus ponens (5,090 applications), reductio ad absurdum (5,084), disjunction elimination (5,060), modus tollens (5,001), disjunctive syllogism (4,925), hypothetical syllogism (2,159) and constructive dilemma (681). The stated label agrees with the checker on 6,972 of the 7,000 instances (99.60\%).}

\section{Details on Human Participants}
\label{app:human}

\revision{18 anonymous participants each answer a subset of the questions. This is because deductive reasoning questions, especially those at large proof sizes, are cognitively demanding and time-consuming; it is impractical to expect humans to complete 1050 questions. Participants were assigned one of six fixed question sets, three participants per set. Each set contains seven questions, one at each proof size, so every participant answered exactly seven questions and the set of responses covers each proof size equally. To ensure fairness, both models and participants are provided similar prompts. To ensure that participants understand the requirements of the task, a simple verification question is added. If they answer incorrectly, their subsequent responses are voided.}

\revision{Across the resulting 126 responses the human average is 73.0\% (Wilson 95\% CI [64.7, 80.0]; participant-level bootstrap 95\% CI [64.3, 81.7]; participant-level standard deviation 19.5\%). Per-participant scores, sorted, are 100, 100, 100, 100, 85.7, 85.7, 85.7, 71.4, 71.4, 71.4, 71.4, 57.1, 57.1, 57.1, 57.1, 57.1, 42.9 and 42.9. Accuracy by proof size is 77.8, 55.6, 77.8, 66.7, 77.8, 72.2 and 83.3 for proof sizes 1--7, with 18 responses at each proof size. Unlike every model we evaluate, human accuracy shows no measurable degradation with proof size, though with 18 responses at each the intervals are too wide to support a conclusion either way.}

\begin{figure}[H]
	\centering
	\subfloat{{\includegraphics[height=4cm, keepaspectratio]{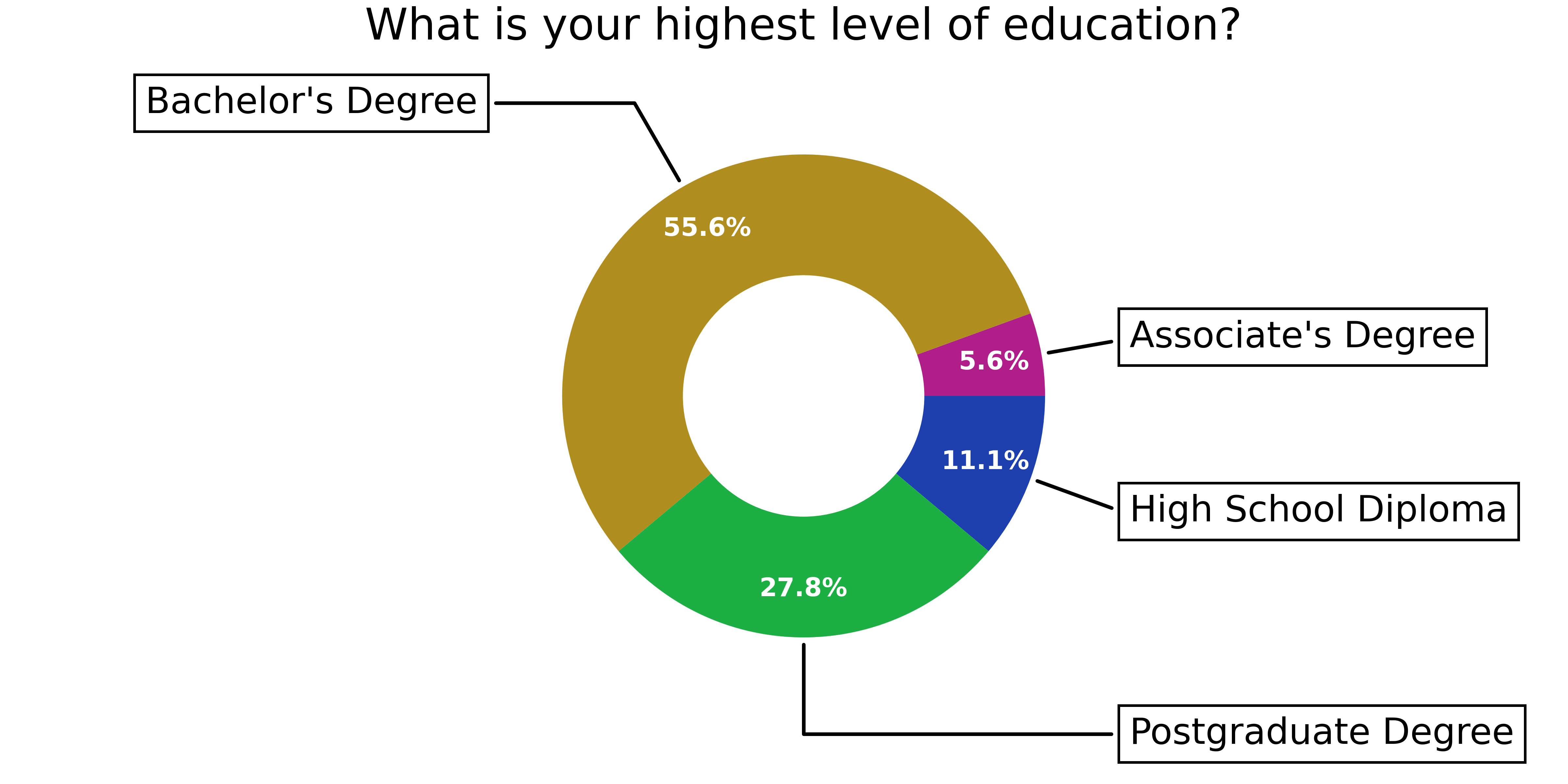} }}%
	\quad
	\subfloat{{\includegraphics[height=4cm, keepaspectratio]{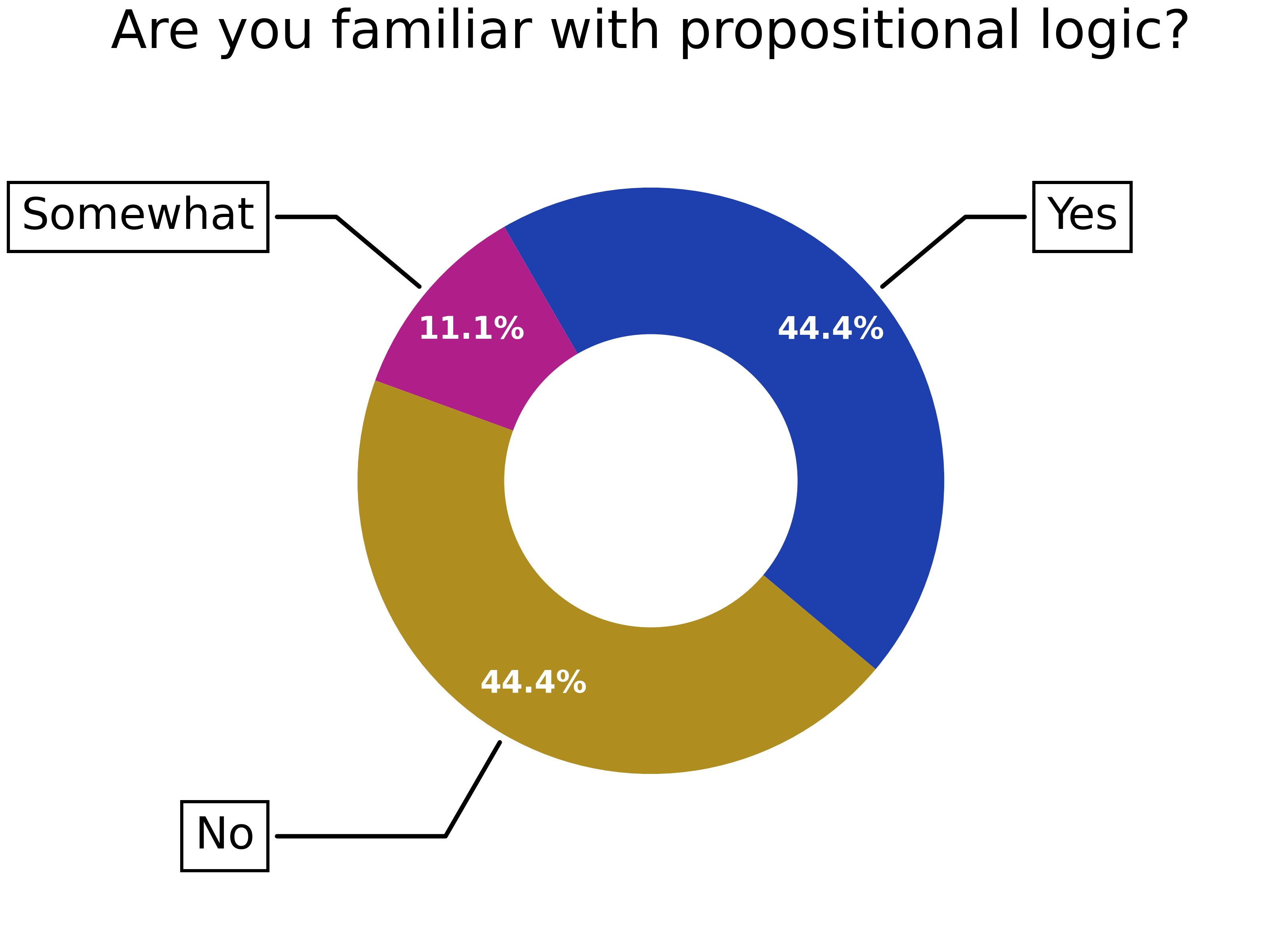} }}
	\caption{\revision{Participants' highest level of education and familiarity with proposition logic.}}%
	\label{fig:human}%
\end{figure}

\revision{Participants are recruited from Amazon Mechanical Turk \citep{amazon2005} and are paid \$24 per hour. Participation is entirely voluntary, and the survey posed no foreseeable risks to participants. As reflected in Figure \ref{fig:human}, to create a sample representative of the human average, the participants possess a diverse range of educational qualifications and familiarity with propositional logic. The highest observed participant accuracy (100.0\%) was achieved by four of the eighteen participants, each on seven questions. Since every JustLogic instance has a unique, logically determined correct answer, a sufficiently careful reasoner can in principle achieve perfect accuracy. A 7/7 score has a Wilson 95\% interval of [64.6, 100.0], so we report the 100.0\% figure as the highest accuracy observed in this study rather than as an estimate of an upper bound on human performance.}

\section{Additional Model Evaluations}
\label{app:model_evals}

\subsection{Additional Evaluations on Various Prompting Techniques}
\label{app:prompt_performance}

\revision{While reasoning models do not require specific prompting techniques due to their reasoning-specific training, non-reasoning models observe significant deltas in accuracy based on the choice of prompts. Thus, we evaluate \revision{the four non-reasoning models} on additional prompting techniques: (i) the self-consistency decoding (SC) \citep{wang2022self}, where the answer is derived through majority voting over 5 sampled paths, and (ii) the tree-of-thought (ToT) framework \citep{yao2023tree}, where each step generates 3 candidates and ultimately chooses 1; the maximum steps allowed is $\text{proof size} + 2$, but the model may terminate the search earlier. Finally, we also test (iii) a CoT prompt that does not mention propositional logic. Explicit mentions of technical terms in propositional logic, e.g. reductio ad absurdum, may hinder the reasoning ability of models that are less familiar with them. This prompt tests the aforementioned hypothesis. \revision{Table \ref{table:results} reports the results.}}

\begin{table}[H]
	\def\arraystretch{1.2}
	\centering
		\caption{\revision{Accuracy (\%) of non-reasoning models on JustLogic under six prompting techniques. The best technique for each model is in bold.}}
	\label{table:results}
    \color{revisioncolor}
	\resizebox{0.85\textwidth}{!}{%
	\begin{tabular}{lcccccc}
		\thickhline
		& \textbf{0-shot} & \textbf{Few-shot} & \textbf{CoT} & \textbf{SC-CoT} & \textbf{ToT} & \textbf{CoT (w/o prop. logic)} \\ \thickhline
		Llama3-8B-Instruct & 49.8 & 41.8 & \textbf{57.8} & 54.6 & 38.6 & 54.0 \\
        Llama3-70B-Instruct & 53.1 & 57.8 & \textbf{64.6} & 58.6 & 60.6 & 58.3 \\
		GPT-4o-mini & 53.0 & \textbf{54.7} & 51.8 & 50.3 & 48.6 & 50.0 \\
        GPT-4o & 53.8 & 58.3 & 65.6 & 67.1 & \textbf{71.4} & 67.4 \\
        \thickhline
	\end{tabular}%
	}
\end{table}

\revision{The relative performance of the prompting techniques is heterogeneous across models. However, besides GPT-4o, we find that prompting techniques that are more expensive than vanilla CoT offer little to no performance advantage. Self-consistency CoT achieves similar performance to CoT; the former may require significantly higher sampled paths to reap its benefits. Tree-of-thought is too complex for most models to utilize, often hallucinating across prompts and failing to break down the problem into coherent steps. Lastly, we find that the explicit mention of propositional logic in the prompt is generally helpful towards model performance.}

\subsection{Accuracy by Answer Class}
\label{app:per_class}

\revision{Table \ref{table:per_class} reports accuracy within each gold class, together with the share of items each model assigns to Uncertain. Every model is strongest on Uncertain, and the weaker the model the larger that advantage: the gap between accuracy on Uncertain and the mean of accuracy on True and False ranges from 10.6 points for GPT-5.5 to 65.3 points for GPT-4o-mini, which reaches 94.3\% on Uncertain but only 25.3\% on True. The predicted-label distributions are skewed the same way: GPT-4o-mini answers Uncertain on 67.7\% of items against a gold rate of 34.9\%, and OpenAI o1-mini, Llama3-70B and GPT-4o all exceed 55\%. Uncertain is also the class the construction artifact of Appendix \ref{app:artifacts} identifies almost perfectly, so the two observations are related.}

\begin{table}[H]
	\centering
	\caption{\revision{Accuracy within each gold class, and the share of items assigned to Uncertain. On the full test set the gold classes are 32.8\% True, 32.3\% False and 34.9\% Uncertain.}}
	\label{table:per_class}
	\revision{
	\begin{tabular}{lccccc}
		\thickhline
		& \textbf{Overall} & \textbf{True} & \textbf{False} & \textbf{Uncertain} & \begin{tabular}[c]{@{}c@{}}\textbf{Predicted}\\ \textbf{Uncertain (\%)}\end{tabular} \\ \thickhline
		GPT-5.5 & 88.8 & 86.9 & 83.2 & 95.6 & 43.0 \\
		Gemini 2.5 Flash & 82.0 & 76.5 & 74.9 & 93.7 & 45.4 \\
		DeepSeek R1 & 80.9 & 78.6 & 69.3 & 95.5 & 43.4 \\
		OpenAI o1 & 72.9 & 63.0 & 63.6 & 95.2 & 50.0 \\
		GPT-4o & 65.6 & 56.7 & 45.1 & 92.9 & 55.3 \\
		Llama3-70B & 64.6 & 54.9 & 40.1 & 96.2 & 56.3 \\
		OpenAI o1-mini & 62.0 & 48.4 & 46.5 & 93.6 & 59.1 \\
		Qwen 14B (R1 Distill) & 61.7 & 49.2 & 46.5 & 91.8 & 54.9 \\
		Llama3-8B & 57.8 & 58.1 & 46.6 & 67.8 & 35.0 \\
		GPT-4o-mini & 51.8 & \textbf{25.3} & 32.7 & \textbf{94.3} & \textbf{67.7} \\ \thickhline
	\end{tabular}}
\end{table}

\subsection{Understanding the Performance of OpenAI o1 vs. GPT-5.5}
\label{app:o1}

\begin{figure}[H]
	\centering
	\includegraphics[width=8cm, keepaspectratio]{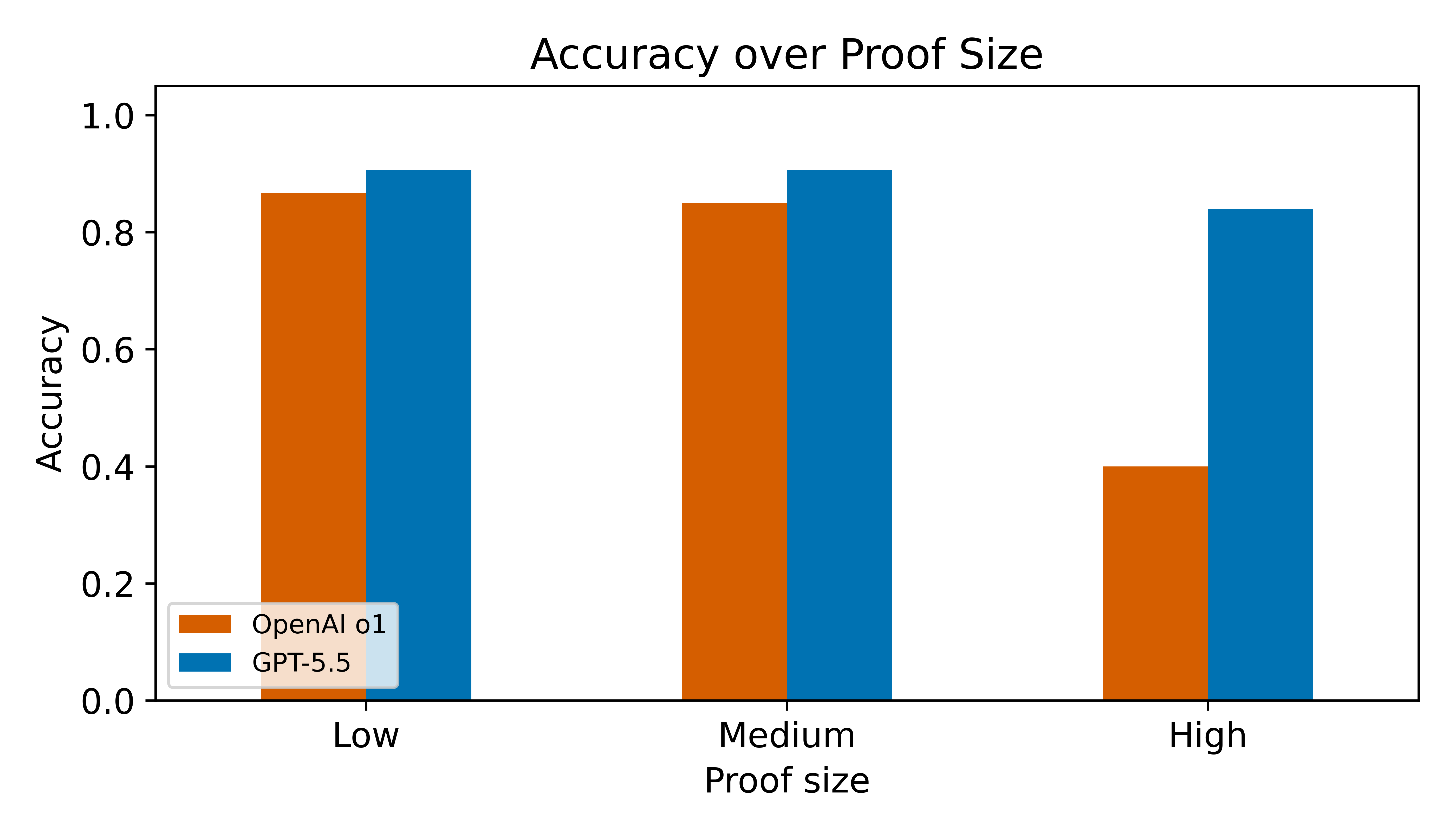}
	\caption{How proof size affects accuracy for OpenAI o1 and GPT-5.5.}%
	\label{fig:acc_depth_o1}%
\end{figure}

\revision{OpenAI o1 obtains 72.9\% accuracy on its 70-item evaluation subset, compared with 87.1\% for GPT-5.5 on the same items; this difference does not survive Holm correction across the 55 pairwise model comparisons. We qualitatively analysed o1's responses, which are available in our repository. In the inspected responses, o1 often produces shorter answers at proof sizes 5--7 than at proof sizes 3--4, and it answers ``Uncertain'' on 90\% of its proof-size-7 items. Figure \ref{fig:acc_depth_o1} reports accuracy across proof-size groups: at proof sizes 6--7, o1 achieves 0.40 accuracy, compared with 0.84 for GPT-5.5.}

\revision{These response patterns accompany o1's lower accuracy at larger proof sizes, whereas GPT-5.5 maintains higher observed accuracy across the evaluated proof-size groups. Performance as a function of longest inference-chain length is reported separately in Appendix \ref{app:size_vs_chain}. Importantly, this case study reflects JustLogic's ability to flexibly probe models at different proof-size ranges.}

\subsection{Model Evaluations on a Common Subset}
\label{app:common_subset}

\revision{The evaluation subsets are nested, because each is formed by taking the first $n$ instances at every proof size: the 70 items used for OpenAI o1 are contained in the 350 used for the reasoning models, which are contained in the full 1,050, and the three 350-item runs use an identical set of items. Every model can therefore be re-scored on a common subset without any additional evaluation, and every pair of models can be compared with an exact paired McNemar test. Table \ref{table:common70} re-scores all models on the 70 items OpenAI o1 was run on.}

\begin{table}[H]
	\centering
	\caption{\revision{All models re-scored on the common 70-item subset, with Wilson 95\% intervals.}}
	\label{table:common70}
	\revision{
	\begin{tabular}{lcc}
		\thickhline
		& \textbf{Accuracy} & \textbf{95\% CI} \\ \thickhline
		GPT-5.5 & \textbf{87.1} & [77.3, 93.1] \\
		DeepSeek R1 & 85.7 & [75.7, 92.1] \\
		Gemini 2.5 Flash & 81.4 & [70.8, 88.8] \\
		OpenAI o1 & 72.9 & [61.5, 81.9] \\
		GPT-4o & 64.3 & [52.6, 74.5] \\
		Qwen 14B (R1 Distill) & 64.3 & [52.6, 74.5] \\
		Llama3-70B & 62.9 & [51.1, 73.2] \\
		OpenAI o1-mini & 58.6 & [46.9, 69.4] \\
		Llama3-8B & 52.9 & [41.3, 64.1] \\
		GPT-4o-mini & 51.4 & [40.0, 62.8] \\ \thickhline
	\end{tabular}}
\end{table}

\revision{The subset also affects which differences are significant: with Holm correction, 35 of the 55 model pairs are significant when each pair is compared on its largest shared subset, but only 15 of 55 on these 70 items.}

\subsection{Futureproofing JustLogic}
\label{app:futureproof}

\revision{As LLMs improve, we expect their performance on JustLogic to rise, which motivates generating instances with larger proof structures. One way is to increase the proof size: specifically, we extended JustLogic to incorporate questions of very large proof size (8 to 11), and evaluated them on the current SOTA reasoning and non-reasoning models, i.e. DeepSeek R1 and GPT-4o, using CoT prompt.}

\revision{Table \ref{table:futureproof_results} reports accuracy across four JustLogic proof-size ranges, together with results on LogiQA 2.0 and FOLIO. For both evaluated models, accuracy decreases as JustLogic proof size increases, reaching 53.0\% for GPT-4o and 65.0\% for DeepSeek R1 at proof sizes 8--11. The LogiQA 2.0 and FOLIO results provide cross-benchmark reference points under their respective logical languages and task formats.}

\begin{table}[H]
	\def\arraystretch{1.2}
	\centering
	\caption{Model accuracy across JustLogic proof-size ranges and on LogiQA 2.0 and FOLIO.}
	\label{table:futureproof_results}
	\color{revisioncolor}
	\begin{tabular*}{0.75\textwidth}{@{\extracolsep{\fill}}lcccccc@{}}
		\thickhline
		& \multicolumn{4}{c}{JustLogic (proof size)}
		& \multirow{2}{*}{LogiQA 2.0}
		& \multirow{2}{*}{FOLIO} \\
		\cmidrule(lr){2-5}
		& 1--3 & 4--5 & \textbf{6--7} & \textbf{8--11} & & \\
		\thickhline
		GPT-4o
		& 77.3 & 63.3 & \textbf{57.3} & \textbf{53.0}
		& 64.5 & 76.3 \\
		DeepSeek R1
		& 90.0 & 83.3 & \textbf{68.0} & \textbf{65.0}
		& 87.6 & 86.2 \\
		\thickhline
	\end{tabular*}
\end{table}

\subsection{Extending the Generator with Conjunction}
\label{app:conjunction}

\revision{Conjunction is given the two roles the generator already gives disjunction: it is sampled as a compound proposition, with twelve expression templates, and it is consumed by an argument form, conjunctive modus ponens ($p$, $q$, $(p \wedge q) \rightarrow r \vdash r$), in which two separately stated premises must be assembled into a conjunctive antecedent before modus ponens applies. Conjunction introduction and simplification were not used, since they add a node to the proof tree without adding an inference step. We use the extension to ask whether the choice of connective itself affects difficulty.}

\revision{A conjunctive premise is unavoidably longer, and accuracy tracks paragraph length, so comparing the conjunction arm against the unmodified generator would confound the operator with the extra text. We therefore generated three arms over the same proof trees, matched item by item with identical proof size, query and gold label: a baseline, a length arm padded with logically inert premises to the same length as the conjunction arm, and a conjunction arm in which eligible modus ponens nodes are replaced. Table \ref{table:conjunction} reports 350 matched instances per arm on six models. The operator effect is $-1.0$ points on average, with a pooled 95\% interval of $[-2.9, +0.9]$, and none of the eighteen paired McNemar tests is significant after Holm correction; the signs are inconsistent across models. Adding the operator therefore changes difficulty by at most about one point.}

\begin{table}[H]
	\centering
	\caption{\revision{Accuracy on 350 matched instances per arm; the length effect is baseline $\rightarrow$ length and the operator effect is length $\rightarrow$ conjunction. These instances are generated separately from the test set, so the absolute values are not comparable with Table \ref{table:results_all} --- only the between-arm differences are meaningful.}}
	\label{table:conjunction}
	\revision{
	\begin{tabular}{lccccc}
		\thickhline
		& \textbf{Baseline} & \textbf{Length} & \textbf{Conjunction} & \begin{tabular}[c]{@{}c@{}}\textbf{Length}\\ \textbf{effect}\end{tabular} & \begin{tabular}[c]{@{}c@{}}\textbf{Operator}\\ \textbf{effect}\end{tabular} \\ \thickhline
		Gemini 2.5 Flash & 88.9 & 89.1 & 87.1 & $+0.3$ & $-2.0$ \\
		DeepSeek R1 & 87.1 & 85.7 & 86.6 & $-1.4$ & $+0.9$ \\
		GPT-4o & 71.1 & 70.9 & 70.9 & $-0.3$ & $0.0$ \\
		Llama3-70B & 66.0 & 64.6 & 65.1 & $-1.4$ & $+0.6$ \\
		GPT-4o-mini & 58.9 & 60.3 & 57.4 & $+1.4$ & $-2.9$ \\
		Llama3-8B & 53.1 & 56.3 & 53.7 & $+3.1$ & $-2.6$ \\ \hline
		\textbf{Macro-average} & & & & $\mathbf{+0.3}$ & $\mathbf{-1.0}$ \\ \thickhline
	\end{tabular}}
\end{table}

\section{Qualitative Analysis of Failure Modes}
\label{app:failure-cases}

To identify exactly how JustLogic is challenging for existing LLMs, we conducted a qualitative analysis to identify the 4 major failure modes of various models' responses to JustLogic questions. We primarily rely on analyzing the chain-of-thought responses to investigate how they produced the wrong answers. High-level explanations and abbreviated examples are provided for the failure modes.

\textbf{(1) Logical inconsistency.} Models sometimes produce arguments that rely on premises, implicit or otherwise, that contradicts earlier parts of their chain-of-thought. This causes models to generate incoherent arguments, and ultimately arrive at the wrong conclusion. For example, the following is a truncated response from OpenAI o1:

\begin{tcolorbox}[breakable, colback=gray!10,colframe=gray!10,enlarge top by=0cm, enlarge bottom by=0cm, left=2pt, right=2pt,
fontupper=\fontfamily{pcr}\selectfont\small]
While R implies that researchers are observing data, without specific information, we cannot definitively state R $\rightarrow$ S.

...

Since M $\lor$ R is true and M $\lor$ S is the statement we are evaluating, and we have established M is true if N is true, we can conclude:

\begin{itemize}
    \setlength\itemsep{0em}
    \item M $\lor$ S is true.
\end{itemize}
\end{tcolorbox}

This line of reasoning suggests that $M \lor R$ implies $M \lor S$. For this to be the case, $R \rightarrow S$ must be true. However, earlier in the reasoning chain, o1 says “we cannot definitively state $R \rightarrow S$”. This argument is therefore logically inconsistent.

\textbf{(2) Wrong application of argument forms.} Models sometimes identify the correct argument form to solve the question. However, mistakes are made when applying the form to the specific question context. In the truncated response below by Llama3-70B, disjunctive syllogism is, in fact, the appropriate argument form, but the opposite conclusion should have been made: the statement, pain can be severe and make a person unable to perform normal activities, does not hold.

\begin{tcolorbox}[breakable, colback=gray!10,colframe=gray!10,enlarge top by=0cm, enlarge bottom by=0cm, left=2pt, right=2pt,
fontupper=\fontfamily{pcr}\selectfont\small]
Premise 1: Either condensation is water vapor changing to liquid water or the statement, pain can be severe and make a person unable to perform normal activities, does not hold.

Premise 2: The claim that condensation is water vapor changing to liquid water does not correspond with reality. \\

Using disjunctive syllogism with premises 1 and 2, we conclude that the statement, pain can be severe and make a person unable to perform normal activities, holds.
\end{tcolorbox}

\textbf{(3) Using the wrong argument form.} Sometimes, models attempt to use argument forms that logically cannot be applied to the context. Such mistakes often derail the response, making it entirely incoherent. In the following example of an output by Llama3-70B, modus tollens simply cannot be applied here; reductio ad absurdum should have been used instead.
 
\begin{tcolorbox}[breakable, colback=gray!10,colframe=gray!10,enlarge top by=0cm, enlarge bottom by=0cm, left=2pt, right=2pt,
fontupper=\fontfamily{pcr}\selectfont\small]
Premise 1: One may reasonably assume that if the statement that 'football is a memory-making machine' is incorrect, then fairways are lawns.
Premise 2: Assuming the statement that 'football is a memory-making machine' is incorrect, we know that it is not the case that fairways are lawns. \\

Using modus tollens with premise 1 and 2…
\end{tcolorbox}

\textbf{(4) False interpretation of facts.} Models sometimes misinterpret the natural language facts entirely. This is most clearly seen when models translate sentences into the wrong logical form. While some of these translations seem trivially simple, especially for LLMs, such mistakes are as common as the other failure modes. In the example below, GPT-4o's interpretation of Premise 2 is incorrect: Premise 2 does in fact imply that the statement “most roses grow fairly rapidly” is false because of the word “mistakenly.”

\begin{tcolorbox}[breakable, colback=gray!10,colframe=gray!10,enlarge top by=0cm, enlarge bottom by=0cm, left=2pt, right=2pt,
fontupper=\fontfamily{pcr}\selectfont\small]
Premise 2: Some people mistakenly believe that most roses grow fairly rapidly. \\

We cannot directly use Premise 2 to confirm or deny ($\neg$Q), as it only mentions a mistaken belief rather than the truth value. Therefore, we do not have sufficient information to directly negate (Q).
\end{tcolorbox}

Nonetheless, some sentences are more complex and therefore more prone to false interpretations. In the example below, also by GPT-4o, Premise 2's logical form should be $\neg(A \rightarrow B) \rightarrow \neg C$ instead.

\begin{tcolorbox}[breakable, colback=gray!10,colframe=gray!10,enlarge top by=0cm, enlarge bottom by=0cm, left=2pt, right=2pt,
fontupper=\fontfamily{pcr}\selectfont\small]
Premises:

...

2. ``Given that the claim that if police sergeants receive calls, then good nutrition helps reduce low birth weight, miscarriage and anemia does not reflect reality, it can be inferred that some people mistakenly believe that oil is simply a liquid form of fat.'' \\

From Premise 2: ($\neg$(A$\rightarrow$B))
\end{tcolorbox}

\section{Construction Artifacts and Shortcut Baselines}
\label{app:artifacts}

\revision{Because JustLogic's queries are generated rather than written, the construction itself can leave surface regularities that a model may exploit without reasoning. We fit each baseline on the training split and report it on the 1,050-item test split. Table \ref{table:artifact_baselines} gives the results. A rule that labels an item Uncertain when the query's proposition does not occur in the paragraph, combined with a rule that separates True from False using the negation templates of Table \ref{table:basic_forms}, reaches 71.8\% using surface text alone and 79.6\% when given the generator's symbol table --- higher than seven of the ten models in Table \ref{table:results_all}.}

\begin{table}[H]
	\centering
	\caption{\revision{Shortcut and artifact baselines on the 1,050-item test split.}}
	\label{table:artifact_baselines}
	\revision{
	\begin{tabular}{lcccc}
		\thickhline
		& \textbf{Overall} & \textbf{True} & \textbf{False} & \textbf{Uncertain} \\ \thickhline
		Query-only TF-IDF + logistic regression & 63.2 & 57.3 & 65.8 & 66.5 \\
		TF-IDF (paragraph + query) & 38.4 & 41.3 & 31.0 & 42.5 \\
		Bag-of-words (paragraph + query) & 37.0 & 37.5 & 31.9 & 41.4 \\
		Lexical-overlap rule, surface text only & 71.8 & 64.2 & 64.9 & 85.3 \\
		Lexical-overlap rule, using the generator's symbols & \textbf{79.6} & 71.8 & 71.4 & \textbf{94.6} \\
		Query-template rule (never reads the paragraph) & 66.4 & 66.0 & 71.4 & 62.1 \\ \hline
		Majority class (Uncertain) & 35.0 & 0.0 & 0.0 & 100.0 \\
		Random & 33.3 & 33.3 & 33.3 & 33.3 \\ \thickhline
	\end{tabular}}
\end{table}

\revision{Two independent regularities account for this. The first is lexical overlap: a True or False query is the conclusion or its negation, so its proposition necessarily occurs in the paragraph, whereas an Uncertain query was sampled from outside the argument and does not. The mean longest word $n$-gram shared between query and paragraph is 10.4 words for True and 10.3 for False, but 2.5 for Uncertain, and the fraction of the query's content words covered by the paragraph is 0.85, 0.83 and 0.29 respectively. As a binary Uncertain-versus-not classifier the rule reaches 89.2\% on surface text and 98.1\% with the symbol table. The second is the query template. When the conclusion is a plain statement, which occurs with probability 0.8, the generator returns the sampled sentence verbatim, whereas True and False queries are always rendered through an expression template; only 37.9\% of Uncertain queries carry a template wrapper, against 94.2\% and 99.1\%. A rule that looks only at the query therefore separates Uncertain from the rest with 84.6\% accuracy, which is also what the query-only classifier in Table \ref{table:artifact_baselines} learns. The two are independent: drawing Uncertain queries from the paragraph would defeat the first but not the second.}

\revision{To quantify how much of each model's accuracy this accounts for, we re-scored every model on two subsets on which the shortcut cannot help: items whose gold label is True or False, on which the lexical-overlap rule always fires, and items the rule itself gets wrong. No new evaluation was required. Table \ref{table:shortcut_free} reports both. The artifact inflates the weaker models substantially and the strongest one very little, so the field widens once it is removed: GPT-5.5's margin over Llama3-70B grows from 24.2 to 34.1 points. Human participants score \emph{higher} where the rule fails, indicating they were not using it.}

\begin{table}[H]
	\centering
	\caption{\revision{Model accuracy re-scored where the shortcut cannot help. $\Delta$ is the change from the headline figure to the ``rule is wrong'' column.}}
	\label{table:shortcut_free}
	\revision{
	\begin{tabular}{lcccc}
		\thickhline
		& \textbf{Table \ref{table:results_all}} & \textbf{True/False only} & \textbf{Where the rule is wrong} & $\Delta$ \\ \thickhline
		GPT-5.5 & 88.8 & 85.1 & 83.2 & $-5.6$ \\
		Gemini 2.5 Flash & 82.0 & 75.7 & 73.8 & $-8.2$ \\
		DeepSeek R1 & 80.9 & 74.2 & 66.2 & $-14.6$ \\
		OpenAI o1 & 72.9 & 63.3 & 68.4 & $-4.4$ \\
		GPT-4o & 65.6 & 51.0 & 55.1 & $-10.5$ \\
		Llama3-70B & 64.6 & 47.6 & 49.1 & $-15.5$ \\
		OpenAI o1-mini & 62.0 & 47.5 & 47.3 & $-14.7$ \\
		Qwen 14B (R1 Distill) & 61.7 & 47.9 & 47.3 & $-14.4$ \\
		Llama3-8B & 57.8 & 52.4 & 55.1 & $-2.7$ \\
		GPT-4o-mini & 51.8 & 29.0 & 43.5 & $-8.4$ \\ \hline
		\textbf{Human} & \textbf{73.0} & \textbf{67.9} & \textbf{77.8} & $\mathbf{+4.8}$ \\ \thickhline
	\end{tabular}}
\end{table}

\revision{We also ran two premise controls directly on models, over all 1,050 test items. Shuffling the premises within an instance leaves the label valid, and accuracy barely moves: $+2.8$ points for GPT-4o-mini and $-3.3$ for Llama3-8B. Swapping the paragraph across instances destroys the entailment, and the reference point is then the Uncertain base rate of 35.0\%, not 33.3\%, because a model reasoning correctly should answer Uncertain throughout and would be scored against the original labels. GPT-4o-mini scores 34.95\% and answers Uncertain on every item; Llama3-8B scores 36.95\%. Once the paragraph does not match, the label cannot be recovered. We note that GPT-4o-mini already over-predicts Uncertain (Appendix \ref{app:per_class}), so its behaviour here mixes correct reasoning with a known bias; Llama3-8B, whose bias is milder, is the more informative of the two.}

\section{Future Works}

JustLogic can be scaled to incorporate more question types related to logical reasoning, such as multiple-choice questions, identifying missing premises in arguments, identifying logical fallacies in arguments, and natural language sentence to formal logic translation. \cite{liu2023logicot} provides a comprehensive taxonomy. JustLogic's program can be adapted to accommodate each question type while maintaining its key advantages. By measuring deductive reasoning across multiple modalities using a single dataset construction method, JustLogic can provide more comprehensive and controlled evaluations and error analysis.

JustLogic is currently an English, text-based dataset. However, importantly, it has enormous potential to be multilingual and multimodal. To make the dataset multilingual, researchers need only replace the natural language sentences used in Step 2 of the dataset construction process, i.e., replace the list of logical form expression templates and GenericsKB database. Otherwise, the dataset construction process is generalizable to any language. It is also entirely feasible to make the dataset multimodal. As a straightforward example, symbols can be used to reference other modalities, rather than just being replaced by text. For example, “If x, then y.” can be replaced by “If the country in the map is shaded red, then it is currently in a recession.”. JustLogic’s extensibility and flexibility create immense, exciting opportunities for future research.

\end{document}